\documentclass[twoside,11pt]{article}

\usepackage{jmlr2e}
\usepackage{algpseudocode,multirow, float,bm,amsmath,xcolor,colortbl}
\newcommand{\md}[1]{\mathbb{#1}}

\newcommand{\thickhline}{%
	\noalign {\ifnum 0=`}\fi \hrule height 1pt
	\futurelet \reserved@a \@xhline
}
\renewcommand{\P}{\mathbb{P}}

\newtheorem{cor}{Corollary}
\newtheorem{theo}{Theorem}
\newtheorem{lem}{Lemma}
\newtheorem{defi}{Definition}
\newtheorem{assumption}{Assumption}
\newtheorem{prop}{Proposition}
\renewcommand{\P}{\mathbb{P}}

\DeclareMathOperator\diag{Diag}
\DeclareMathOperator\rank{Rank}
\DeclareMathOperator\poly{poly}

\makeatletter
\newcommand{\mytag}[2]{%
	\text{#1}%
	\@bsphack
	\begingroup
	\@onelevel@sanitize\@currentlabelname
	\edef\@currentlabelname{%
		\expandafter\strip@period\@currentlabelname\relax.\relax\@@@%
	}%
	\protected@write\@auxout{}{%
		\string\newlabel{#2}{%
			{#1}%
			{\thepage}%
			{\@currentlabelname}%
			{\@currentHref}{}%
		}%
	}%
	\endgroup
	\@esphack
}
\makeatother

\usepackage{algorithm}

\usepackage{lastpage}
\jmlrheading{21}{2020}{1-\pageref{LastPage}}{6/19; Revised
	10/20}{11/20}{19-471}{Dimitris Bertsimas and Michael Lingzhi Li}
\ShortHeadings{Fast Exact Matrix Completion}{Bertsimas and Li}

\firstpageno{1}

\begin{document}

\title{Fast Exact Matrix Completion: A Unified Optimization Framework for Matrix Completion}

\author{\name Dimitris Bertsimas \email dbertsim@mit.edu \\
       \addr Sloan School of Management and Operations Research Center\\
       Massachusetts Institute of Technology\\
       Cambridge, MA 02139, USA
       \AND
       \name Michael Lingzhi Li \email mlli@mit.edu \\
       \addr Operations Research Center\\
       Massachusetts Institute of Technology\\
       Cambridge, MA 02139, USA}

\editor{Tong Zhang}

\maketitle

\begin{abstract}
We formulate   the problem of matrix completion  with and without side information as a non-convex optimization problem. 
We design fastImpute based on non-convex gradient descent and   show it converges to a global minimum that is guaranteed to 
recover  closely the underlying matrix while it scales to matrices of sizes beyond $10^5 \times 10^5$. We report experiments on both synthetic and real-world datasets that show fastImpute is competitive in both the accuracy of the matrix recovered and the time needed across all cases. Furthermore, when a high number of entries are missing, fastImpute is over $75\%$ lower in MAPE and $15$ times  faster than current state-of-the-art matrix completion methods in both the case with side information and without.
\end{abstract}


\section{Introduction\label{sec:Introduction}}
Low-rank matrix completion is one of the most studied problems after its successful application in the Netflix Competition. It  has been used in 
 computer vision (\cite{candes2010matrix}), signal processing (\cite{ji2010robust}), and control theory (\cite{boyd1994linear}) to generate a completed matrix from partially observed entries,
 among several other areas. Given a data matrix $\bm{A} \in \md{R}^{n\times m}$ , the low-rank assumption assumes that rank$({\bm A})$ is small - in other words 
there are only a few, but still  unknown, common linear factors that  affect  $A_{ij}$. 

In recent years, as noted by \cite{nazarov2018sparse},  there has been a rise in interest for \emph{inductive matrix completion}, where the common linear factors are chosen from a set of given vectors in the form of side information.

In this paper, we present an optimization based approach that improves upon the state of the art in matrix completion with and without side information. We next  review the literature in both the general matrix completion  and  the inductive matrix completion areas.
\subsection*{Literature}
\subsubsection*{General Matrix Completion}
Matrix completion has been applied successfully for many tasks, including recommender systems \cite{koren2009matrix}, social network analysis \cite{chiang2014prediction} and clustering \cite{chen2014clustering}. After \cite{candes2009exact} first proved a theoretical guarantee for the retrieval of the exact matrix under the nuclear norm convex relaxation, a lot of methods have focused on the nuclear norm problem (see \cite{SoftImpute}, \cite{beck2009fast}, \cite{jain2010guaranteed}, and \cite{tanner2013normalized} for examples). Alternative methods include alternating projections by \cite{recht2013parallel} and Grassmann manifold optimization by \cite{keshavan2009matrix}. There has also been work where the uniform distributional assumptions required by the theoretical guarantees are violated, such as \cite{negahban2012restricted} and \cite{chen2014coherent}. 

Despite the non-convexity of the problem, many gradient-descent based approaches have also been proposed. Many works, including \cite{koren2009matrix}, \cite{jain2015fast}, \cite{zheng2016convergence} and \cite{jin2016provable}, utilize the Burer-Monteiro factorization ($\bm{A}=\bm{U}\bm{V}^T$, where $\bm{U} \in \md{R}^{n \times k}$ and $\bm{V}\in \md{R}^{m \times k}$) and conduct various forms (projected, stochastic, lifted, etc) of gradient descent on $\bm{U}$ and $\bm{V}$. Numerical experiments suggest that such algorithms can often converge to the global optimal solution despite their local nature.

Recently, a line of work, including \cite{chen2015fast}, \cite{ge2016matrix}, and \cite{ma2019implicit}, investigated the reasons behind this uncanny efficiency of gradient descent on matrix completion. Collectively, they showed that for the case where $\bm{A}$ is positive semi-definite (PSD), gradient descent on the symmetric factorization $\bm{A}=\bm{U}\bm{U}^T$ can converge linearly to the global optimum under restricted isometry conditions. 

This work has also been reproduced in the non-PSD case considered here ($\bm{A}=\bm{U}\bm{V}^T$) if special regularization terms are added (see for example \cite{sun2016guaranteed}).

Our work differs from previous work in one significant way: After writing $\bm{A}=\bm{U}\bm{V}^T$, instead of performing gradient descent on both $\bm{U}$ and $\bm{V}$, we derive $\bm{U}$ as a function of $\bm{V}$, $g(\bm{V})$, and then directly perform gradient updates for $\bm{V}$, with a projection to a fixed-norm hypersphere $\|\bm{V}\|_2=1$ to ensure scaling invariance. 

Using this formulation, we are able to prove convergence guarantees without special regularization terms on the objective. Numerical experiments show that we are able to retrieve the matrix  faster  than the top performing methods discussed here.

\subsubsection*{Inductive Matrix Completion}
Interest in inductive matrix completion intensified after \cite{xuspeedup} showed that given predictive side information, one only needs $O(\log n)$ samples to retrieve the full matrix. 
Thus, most of this work (see \cite{xuspeedup}, \cite{jain2013provable}, \cite{farhat2013genomic},  \cite{natarajan2014inductive}) have focused on the case in which the side information is assumed to be perfectly predictive so that the theoretical bound of $O(\log n)$ sample complexity \cite{xuspeedup} can be achieved. 
 \cite{chiang2015matrix} explored the case in which the side information is corrupted with noise, while \cite{shah2017matrix} and \cite{si2016goal} incorporated nonlinear combination of factors into the side information. As pointed out by a recent article \cite{nazarov2018sparse}, there is a considerable lack of effort to introduce sparsity into inductive matrix completion, with \cite{lu2016sparse}, \cite{soni2016noisy}, \cite{nazarov2018sparse} as examples.  \cite{bertsimas2018interpretable}  introduced a convex binary formulation of the sparse inductive matrix completion problem, and constructed randomized algorithms for scaling.
 
 Our work differs from the work  in \cite{lu2016sparse}, \cite{soni2016noisy}, and \cite{nazarov2018sparse}  as it does not consider the heuristic convex relaxation of sparsity in the nuclear norm, but rather the exact sparse problem. \cite{bertsimas2018interpretable} consider the problem where the low-rank factorization needs to be selected from features in the side information, while we allow the factorization to be any linear combination of the features in the side information. 
  This greatly increases the flexibility of the algorithm, provides higher  modeling power, and leads to stronger matrix recovery.

\subsection*{Contributions and Structure}
Our contributions in this paper are as follows:
\begin{enumerate}
	\item We reformulate the low-rank matrix completion problem, both with side information and without,  as a separable optimization problem. We show that the general matrix completion problem is in fact a special case of the matrix completion problem with side information.
	\item We propose a novel algorithm using projected gradient descent and Nesterov's accelerated gradient to solve the reformulated matrix completion problem. We prove that the algorithm is guaranteed to converge to the optimal solution under mild conditions. 
	\item We present computational results on both synthetic and real-world datasets that shows the algorithm outperforms current state-of-the-art methods in scalability and accuracy
	 with and without side information. In particular, for the general matrix completion problem, we show the algorithm is about $15$ times  faster than the fastest algorithm available, while achieving on average a $75\%$ decrease in error of retrieval on synthetic datasets. 
\end{enumerate}
The structure of the paper is as follows.
In Section  \ref{sec:mc_separable}, we introduce the separable reformulation of the low-rank matrix completion problem.
In Section \ref{sec:mc_graddescent}, we introduce the base projected gradient descent method, projImpute.
In Section \ref{sec:mc_pgdside}, we introduce fastImpute, the stochastic version of projImpute designed for use with side information. We report on its computational complexity, derive theoretical results, and compare it with other algorithms.
in Section \ref{sec:mc_computeside}, we report results on both synthetic and real-world datasets with side information using fastImpute and compare its performance with other algorithms.
In Section \ref{sec:mc_computegeneral}, we show how the calculations can be further simplified when there is no side information. We report results on both synthetic and real-world datasets without side information using fastImpute and compare its performance with other algorithms.
In Section \ref{sec:mc_conclusion}, we provide our conclusions. 

\section{Reformulation of Matrix Completion}
\label{sec:mc_separable}
The classical  matrix completion problem considers a matrix $\bm{A} \in \md{R}^{n\times m}$ in which $\Omega=\{(i,j) \; |\; A_{ij}\text{ is known}\}$ is the set of known values. We aim to recover a matrix $\bm{X}=\bm{U}\bm{V}^T$ of  rank $k$ that minimizes the 
distance between  $\bm{X}$ and $\bm{A}$ on the known entries $\bm{A}$:
\begin{equation*}
\min_{\bm{X}} \frac{1}{nm} \sum_{(i,j) \in \Omega} (X_{ij}-A_{ij})^2  \quad \text{subject to} \quad \rank(\bm{X})=k.
\end{equation*}
The problem we consider here is that for every column  $j=1,\ldots, m$, we have a given  $p$-dimensional feature vector $\bm{B}_j$  with $p\geq k$  that contains 
the information we have on column $j$. In the Netflix  example, column $j$ corresponds to movie $j$, and thus the feature vector $\bm{B}_j$  includes information 
about the $j$th movie: Budget,  Box Office revenue, IMDB rating, etc. We represent all this side information with a matrix $\bm{B} \in \md{R}^{m \times p}$. 
Given side data $\bm{B}$ we postulate that $\bm{X}=\bm{U}\bm{S}^T\bm{B}^T$, where $\bm{U} \in \md{R}^{n \times k}$ is the matrix of feature exposures, and 
\begin{equation*}
\bm{S}=\begin{pmatrix}
s_{11} & s_{12} & \cdots & s_{1k}\\
s_{21} & s_{22} & \cdots & s_{2k}\\
\vdots & \vdots & \ddots & \vdots\\
s_{p1} & s_{p2} & \cdots & s_{pk}\\
\end{pmatrix} \in \md{R}^{p \times k}.
\end{equation*}
Then, the matrix completion problem with side data $\bm{B}$ can be written as:
\begin{equation*}
\min_{\bm{S}} ~\min_{\bm{U}}\frac{1}{nm} \sum_{(i,j) \in \Omega} (X_{ij}-A_{ij})^2  \quad \text{subject to} \quad \bm{X}=\bm{U}\bm{S}^T\bm{B}^T,\; \|\bm{S}\|_2=1.
\end{equation*}
The norm constraint on $\bm{S}$ reduces the space of feasible solutions without loss of generality, as the problem is invariant under the transformation $\bm{U} \to \bm{U}\bm{D}^{-1}$ and $\bm{S}^T \to \bm{D}\bm{S}^T$ for any diagonal matrix $\bm{D} \in \md{R}^{k\times k}$ that is invertible.  Throughout the paper  $\|\cdot\|_2$ is   the Frobenius norm. We note that since $\bm{S}$ is a $p\times k$ matrix, the rank of matrix $\bm{X}$ is indeed $k$. Further,we note that if $p=m$, and $\bm{B}=\bm{I}_m$, then the problem is reduced back to the general low-rank matrix completion problem with no feature information.

Similar to linear  regression and for robustness purposes as shown in \cite{bertsimas2018characterization}, we address in this paper  the problem with a Tikhonov regularization term. Specifically, the matrix completion problem with side information and regularization 
we address is
\begin{equation}
\min_{\|\bm{S}\|_2=1} ~\min_{\bm{U}} \frac{1}{nm}\left(\sum_{(i,j) \in \Omega} (X_{ij}-A_{ij})^2 + \frac{1}{\gamma}\|\bm{U}\|_2^2\right) \quad \text{subject to} \quad \bm{X}=\bm{U}\bm{S}^T\bm{B}^T \label{mainproblem},
\end{equation}
where  $\gamma>0$ is a given parameter that controls the strength of the regularization term. We can reformulate Problem (\ref{mainproblem}) as followed:
\begin{prop}
	\label{prop:maintheo}
	Problem (\ref{mainproblem}) can be reformulated as a separable optimization problem:
	\begin{align}
	\min_{\|\bm{S}\|_2=1} ~c(\bm{S})&=\frac{1}{nm} \sum_{i=1}^n \overline{\bm{a}}_i
	\left(\bm{I}_m-\bm{V}\left(\frac{\bm{I}_k}{\gamma}+\bm{V}^T\bm{W}_i\bm{V}\right)^{-1}\bm{V}^T\right)\overline{\bm{a}}_i^T ,\label{eq:mc_central}
	\end{align}
	where $\bm{V}=\bm{B}\bm{S}$, $\bm{W}_1,\cdots, \bm{W}_n\in \md{R}^{m\times m}$
	are   diagonal  matrices:
	\[(\bm{W}_{i})_{jj}=\begin{cases}
	1, & (i,j) \in \Omega, \vspace{3pt} \\
	0, & (i,j) \not \in \Omega,
	\end{cases}\]
	and $\overline{\bm{a}}_i=\bm{a}_i\bm{W}_i$, $i=1\ldots, n$, so that $\overline{\bm{a}}_i \in \md{R}^{1 \times m}$ is the $i$th row of $\bm{A}$ with unknown entries taken to be 0.
\end{prop}
\proof
	With the diagonal projection matrices $\bm{W}_i$ defined above, we can rewrite the sum in (\ref{mainproblem}) over known entries of $\bm{A}$, $ \sum_{(i,j) \in \Omega} (X_{ij}-A_{ij})^2$, as a sum over the rows of $\bm{A}$:
	\begin{equation*}
	\sum_{i=1}^n \|(\bm{x}_i-\bm{a}_i)\bm{W}_i\|_2^2,
	\end{equation*}
	where $\bm{x}_i$ is the $i$th row of $\bm{X}$. Using $\bm{X}=\bm{U}\bm{S}^T\bm{B}^T$, then  $\bm{x}_i=\bm{u}_i\bm{S}^T\bm{B}^T$ where  $\bm{u}_i$ is the $i$th row of $\bm{U}$. Moreover, 
	$$\|\bm{U}\|_2^2=\sum_{i=1}^n \|\bm{u}_i\|_2^2.$$
	Then, Problem (\ref{mainproblem}) 
	becomes:
	$$
	\min_{\|\bm{S}\|_2=1} ~\min_{\bm{U}} \frac{1}{nm}\left( \sum_{i=1}^n \left(\|(\bm{u}_i\bm{S}^T\bm{B}^T-\bm{a}_i)\bm{W}_i\|_2^2 + \frac{1}{\gamma}\|\bm{u}_i\|_2^2\right) \right).
	$$
	We then notice that within the sum $\sum_{i=1}^n$  each row of $\bm{U}$ can be optimized separately, leading to:
	\begin{equation}
	\min_{\|\bm{S}\|_2=1} ~\ \frac{1}{nm}\left( \sum_{i=1}^n \min_{\bm{u}_i}\left(\|(\bm{u}_i\bm{S}^T\bm{B}^T-\bm{a}_i)\bm{W}_i\|_2^2 + \frac{1}{\gamma}\|\bm{u}_i\|_2^2\right) \right).\label{intermediate}
	\end{equation}
	The inner optimization problem $\displaystyle \min_{\bm{u}_i}\|(\bm{u}_i\bm{S}^T\bm{B}^T-\bm{a}_i)\bm{W}_i\|_2^2 + \frac{1}{\gamma}\|\bm{u}_i\|_2^2$ can be solved in closed form given $\bm{S}$, as it is a weighted linear regression problem with Tikhonov regularization. The closed form solution is:
	\begin{equation}
	\bm{a}_i\bm{W}_i(\bm{I}_m+\gamma \bm{W}_i\bm{B}\bm{S}\bm{S}^T\bm{B}^T\bm{W}_i^T)^{-1}\bm{W}_i\bm{a}_i^T= \overline{\bm{a}}_i(\bm{I}_m+\gamma \bm{W}_i\bm{B}\bm{S}\bm{S}^T\bm{B}^T\bm{W}_i)^{-1}\overline{\bm{a}}_i^T\notag  .
	\end{equation}
	So Problem (\ref{intermediate}) can be simplified to:
	$$
	\min_{\|\bm{S}\|_2=1} ~ \frac{1}{nm}\left( \sum_{i=1}^n \overline{\bm{a}}_i(\bm{I}_m+\gamma \bm{W}_i\bm{B}\bm{S}\bm{S}^T\bm{B}^T\bm{W}_i^T)^{-1}\overline{\bm{a}}_i^T \right) \notag .
	$$
	Then, let us define $\bm{V}=\bm{B}\bm{S}$, and write:
	\begin{align*}
	    &\frac{1}{nm}\left( \sum_{i=1}^n \overline{\bm{a}}_i(\bm{I}_m+\gamma \bm{W}_i\bm{B}\bm{S}\bm{S}^T\bm{B}^T\bm{W}_i^T)^{-1}\overline{\bm{a}}_i^T \right)\\& = \frac{1}{nm}\left( \sum_{i=1}^n \overline{\bm{a}}_i(\bm{I}_m+\gamma \bm{W}_i\bm{V}\bm{V}^T\bm{W}_i^T)^{-1}\overline{\bm{a}}_i^T \right)\\&=\frac{1}{nm} \sum_{i=1}^n \overline{\bm{a}}_i
	\left(\bm{I}_m-\bm{V}\left(\frac{\bm{I}_k}{\gamma}+\bm{V}^T\bm{W}_i\bm{V}\right)^{-1}\bm{V}^T\right)\overline{\bm{a}}_i^T,
	\end{align*}
	which is the needed form in (\ref{eq:mc_central}). The second equality comes from the matrix inversion lemma, as derived in \cite{woodbury1950inverting}:
	\begin{lem}
	For matrices $\bm{U} \in \md{R}^{n\times k}$ and $\bm{V} \in \md{R}^{k\times n}$, we have the following equivalence:
	\[(\bm{I}_n+\bm{U}\bm{V})^{-1}=\bm{I}_n-\bm{U}(\bm{I}_k+\bm{V}\bm{U})^{-1}\bm{V}.\]
	\end{lem}
\endproof

\section{A Gradient Descent Algorithm}
\label{sec:mc_graddescent}
In this section, we   describe a gradient descent algorithm to solve the separable optimization problem (\ref{eq:mc_central}), and examine its  properties. First, we  introduce 
\begin{align}
    \alpha_i(\bm{S})=\overline{\bm{a}}_i \bm{\gamma}_i(\bm{S})=\overline{\bm{a}}_i
\left(\bm{I}_m-\bm{V}\left(\frac{\bm{I}_k}{\gamma}+\bm{V}^T\bm{W}_i\bm{V}\right)^{-1}\bm{V}^T\right)\overline{\bm{a}}_i^T \label{eq:mc_alpha_gamma}, ~~i=1,\ldots,n.
\end{align}
Here $\alpha_i(\bm{S})$ is a scalar and $\bm{\gamma}_i(\bm{S})$ is a $m \times 1$ vector. Then the function $c(\bm{S})$ in  (\ref{eq:mc_central}) can be expressed as
$$c(\bm{S})=\frac{1}{nm}\sum_{i =1}^n \alpha_i(\bm{S})=\frac{1}{nm}\sum_{i =1}^n \overline{\bm{a}}_i\bm{\gamma}_i(\bm{S}). $$
Using this notation, we then can calculate the derivative of $c(\bm{S})$:
\begin{lem}
\label{lem:derivresult}
\begin{equation}
\nabla c(\bm{S})=-\frac{2\gamma }{n} \sum_{i=1}^n \bm{B}^T\bm{\gamma}_i(\bm{S})\bm{\gamma}_i(\bm{S})^T\bm{V} .\label{gradformula}
\end{equation}
\end{lem}
\proof
By the standard result of the derivative of the inverse matrix, we have:
\begin{align*}
    \nabla \alpha_i(\bm{S})&=\nabla \overline{\bm{a}}_i\bm{\gamma}_i(\bm{S})\\&=-2\gamma\bm{B}^T\bm{\gamma}_i(\bm{S})\overline{\bm{a}}_i(\bm{I}_m+\gamma \bm{W}_i\bm{V}\bm{V}^T\bm{W}_i^T)^{-1}\bm{B}\bm{S}\\&=-2\gamma \bm{B}^T\bm{\gamma}_i(\bm{S})\bm{\gamma}_i(\bm{S})^T\bm{V}.
\end{align*}
Then, since $c(\bm{S})=\frac{1}{nm}\sum_{i =1}^n \alpha_i(\bm{S})$, we have the required result. 
\endproof

Using Lemma \ref{lem:derivresult}, we apply a projected gradient descent algorithm (as discussed in \cite{bertsekas1997nonlinear}) on the hypersphere $\|\bm{S}\|_2=1$ 
and obtain  in  Algorithm \ref{alg:mc_alg1}. (Note that $\bm{0}^{p\times k}$ denotes a zero matrix of dimension $p\times k$).

\begin{algorithm}[ht] 
	\begin{algorithmic}[1]
		\Procedure{projImpute}{$\bm{A},\bm{B}$,$k$,$\theta$,$t_{max}$}\Comment{$\bm{A} \in \md{R}^{n\times m}$ the masked matrix, $\bm{B} \in \md{R}^{p\times m}$ the feature matrix, $k$ the desired rank, $\theta$ the step size, and $t_{max}$ the number of steps}
		\State $\bm{S}_1\gets \text{random initial matrix with $\|\bm{S}_1\|_2=1$}$ \Comment{Randomized Start}
		\State $\eta_0 \gets \infty $ \Comment{Initialize objective value}
		\State $\eta_1, \bm{G}_1\gets c(\bm{S}_1), \nabla c(\bm{S}_1) $ \Comment{Initialize objective value and gradient}		
		\State $\nabla \tilde{\bm{S}}_{1} \gets \bm{0}^{p \times k}$ \Comment{Initialize accelerated gradient}
		\While{$t<t_{max}$}\Comment{\text{While we have not reached $t_{max}$ iterations}}
		\State $\nabla \tilde{\bm{S}}_{t+1}=\bm{G}_t + \frac{t-1}{t+2}\nabla \tilde{\bm{S}}_{t}$ \Comment{Nesterov accelerated gradient update step} \label{nesterov_step}
		\State $\overline{\nabla \bm{S}_{t+1}} = -\nabla \tilde{\bm{S}}_{t+1} + (\nabla \tilde{\bm{S}}_{t+1} \cdot \bm{S}_t)\bm{S}_t$ \Comment{\text{Project gradient to tangent plane of $\bm{S}_t$}} \label{proj_step}
		\State $\bm{S}_{t+1} \gets  \bm{S}_t \cos \theta + \frac{\overline{\nabla \bm{S}_{t+1}}}{\|\overline{\nabla \bm{S}_{t+1}}\|_2} \sin \theta$ \Comment{\text{Update $\bm{S}_t$ by projected gradient}} \label{update_step}
		\State $\eta_{t+1}, \bm{G}_{t+1}\gets c(\bm{S}_{t+1}), \nabla c(\bm{S}_{t+1}) $ \Comment{Update the new cost and derivative.} \label{calc_cost}
		\State $t \gets t+1$
		\EndWhile
		\State $\bm{S}^*\gets \bm{S}_t$
		\State $i \gets 1$
		\For{$i<n$}
		\State $\bm{a}_i \gets \overline{\bm{a}}_i\bm{B}\bm{S}^{*}(\bm{B}\bm{S}^{*}\bm{W}_i\bm{B}\bm{S}^{*T}\bm{B}^T)^{-1}\bm{S}^{*T}\bm{B}^T$ \label{calc_final} \Comment{Calculate the final $\bm{A}$ matrix} 
		\EndFor
		\State \textbf{return} $\bm{A}$\Comment{Return the filled matrix $\bm{A}$}
		\EndProcedure
	\end{algorithmic}
	\caption{Gradient Descent algorithm for  matrix completion with side information. }
	\label{alg:mc_alg1}
\end{algorithm}

We explain the reasoning behind some key steps of Algorithm  \ref{alg:mc_alg1}.
\paragraph*{Nesterov Step}
This is Step \ref{nesterov_step} of Algorithm \ref{alg:mc_alg1}. We update the gradient with the accelerated formula in \cite{nesterov1983method} by adding the gradient of the current step to $\frac{t-1}{t+2}$ times the previous gradient, which introduces damping in the resulting gradient and enables faster convergence.
\paragraph*{Projection and Update Steps}
This is Step \ref{proj_step} and \ref{update_step} of Algorithm \ref{alg:mc_alg1}. 
There are two reasons why we are optimizing over the $\|\bm{S}\|_2=1$: (a) as explained, rescaling $\bm{S}$ does not change the objective, so the restriction can guarantee a smaller set of feasible solutions;
(b) such restriction   enables us to effectively approximate the objective value, and its derivative using random sampling,   see Theorem \ref{theo:gradapprox}. 

Because we are optimizing on the hypersphere $\|\bm{S}\|_2=1$, our gradient updates need to be projected to the tangent plane of the sphere at the current point $\bm{S}_t$. Thus, we project the raw gradient, $\tilde{\nabla \bm{S}_{t+1}}$ onto the tangent plane to get $\overline{\nabla \bm{S}_{t+1}}$. Then the update step ensures the norm of  the updated $\bm{S}$ is renormalized to  1.

Note that updating the gradient on the hypersphere is a rotation on the great circle formed by  $\bm{S}_t$ and $\overline{\nabla \bm{S}_{t+1}}$, so the update formula is  $\bm{S}_t \cos \theta + \frac{\overline{\nabla \bm{S}_{t+1}}}{\|\overline{\nabla \bm{S}_{t+1}}\|_2} \sin \theta$.

\subsection{Discussion on Computational Complexity}
There are two key computational steps of the algorithm - Step \ref{calc_cost}, where the cost and the derivative is calculated in every gradient update, and Step \ref{calc_final} where the final matrix is calculated. 
We next derive the   asymptotic complexity of such steps.
\begin{prop}
\label{prop:complexproj}
 The computation complexity of Step \ref{calc_cost} in projImpute is $$O\left(|\Omega| \left(p+k^2+\frac{pk+k^3}{m\alpha}\right)\right),$$where $|\Omega|$ is the number of samples given in the original matrix $\bm{A}$ and $\alpha=\frac{|\Omega|}{mn}$ is the percentage not missing. The computational complexity of Step \ref{calc_final} is $$O\left(|\Omega|\left(k^2+\frac{k}{\alpha}+\frac{k^3}{m\alpha}\right)\right).$$
\end{prop}
\proof 
We analyze  Steps \ref{calc_cost}  and \ref{calc_final}   separately. 
\subsubsection*{Computational Complexity of Step \ref{calc_cost}}
Recall we have that:
\begin{align*}
    \alpha_i(\bm{S})=\overline{\bm{a}}_i \bm{\gamma}_i(\bm{S})=\overline{\bm{a}}_i
\left(\bm{I}_m-\bm{V}\left(\frac{\bm{I}_k}{\gamma}+\bm{V}^T\bm{W}_i\bm{V}\right)^{-1}\bm{V}^T\right)\overline{\bm{a}}_i^T, ~~i=1,\ldots,n,
\end{align*}
First let us denote $m_i$ as the number of non-zero entries of $\bm{W}_i$. This is the number of known entries per row. Then  define $\bm{V}_{W_i} \in \md{R}^{m_i\times k}$ as the matrix of $\bm{W}_i\bm{V}$ after removing the all-zero columns, as illustrated below:
\begin{align}
 \bm{W}_i\bm{V}&= \diag(1,0,1,\cdots,0) \times  \begin{pmatrix}
 \text{---} & \bm{v}_1 & \text{---}  \\
 \text{---} & \bm{v}_2 & \text{---} \\ 
 \text{---} & \bm{v}_3 & \text{---} \\
 \vdots & \vdots & \vdots \\
 \text{---} & \bm{v}_k & \text{---} 
 \end{pmatrix}\nonumber\\&=\begin{pmatrix}
 \text{---} & \bm{v}_1 & \text{---}  \\
 \text{---} & \bm{0} & \text{---} \\ 
 \text{---} & \bm{v}_3 & \text{---} \\
 \vdots & \vdots & \vdots \\
 \text{---} & \bm{0} & \text{---} 
 \end{pmatrix} \leadsto \begin{pmatrix}
 \text{---} & \bm{v}_1 & \text{---}  \\
 \text{---} & \bm{v}_3 & \text{---} \\
 \vdots & \vdots & \vdots \\
 \end{pmatrix} :=\bm{V}_{W_i} .\label{simpdefi}
\end{align}
Note that $\bm{V}_{W_i}$ can be created efficiently through subsetting, and its creation does not impact the asymptotic running time. Then similarly, we denote $\bm{a}_{W_i}\in \md{R}^{1 \times m_i}$ as $\bm{a}\bm{W}_i$ with all the zero elements removed. Then we have the following equality, using (\ref{eq:mc_alpha_gamma}):
\begin{align}
    c(\bm{S})&=\frac{1}{nm}\sum_{i =1}^n \overline{\bm{a}}_i\bm{\gamma}_i(\bm{S})
    \nonumber\\&=\frac{1}{nm}\sum_{i =1}^n \overline{\bm{a}}_i\left(\bm{I}_m-\bm{V}\left(\frac{\bm{I}_k}{\gamma}+\bm{V}^T\bm{W}_i\bm{V}\right)^{-1}\bm{V}^T\right)\overline{\bm{a}}_i^T\nonumber \\&=\frac{1}{nm}\sum_{i =1}^n \bm{a}_{W_i}\left(\bm{I}_{m_i}-\bm{V}_{W_i}\left(\frac{\bm{I}_k}{\gamma}+\bm{V}_{W_i}^T\bm{V}_{W_i}\right)^{-1}\bm{V}_{W_i}^T\right)\bm{a}_{W_i}^T  \label{simpprojimpute1} \\&=\frac{1}{nm}\sum_{i =1}^n \bm{a}_{W_i}\bm{a}_{W_i}^T- \bm{a}_{W_i}\bm{V}_{W_i}\left(\frac{\bm{I}_k}{\gamma}+\bm{V}_{W_i}^T\bm{V}_{W_i}\right)^{-1}\bm{V}_{W_i}^T\bm{a}_{W_i}^T.
    \nonumber
\end{align}
The equality (\ref{simpprojimpute1}) follows directly from the definition of $\bm{V}\bm{W}_i$ and $\bm{a}_{W_i}$ as all the removed elements/columns are zero.

Then, the complexity of each term in the sum (\ref{simpprojimpute1}) is $O(m_ik^2+k^3)$. Summing over $n$ terms and noting that:
$$\sum_{i=1}^n nm_i=|\Omega|,$$
we obtain that the full complexity of the objective function is thus $$O(|\Omega| k^2+nk^3)=O\left(|\Omega|\left(k^2 + \frac{1}{m\alpha} k^3\right)\right).$$
Now let us further define:
\begin{equation}
    \bm{\gamma}_{W_i}(\bm{S})=\left(\bm{I}_{m_i}-\bm{V}_{W_i}\left(\frac{\bm{I}_k}{\gamma}+\bm{V}_{W_i}^T\bm{V}_{W_i}\right)^{-1}\bm{V}_{W_i}^T\right)\bm{a}_{W_i}^T, \label{simpgamma}
\end{equation}
and $\bm{B}_{W_i}$ in the same fashion as defined in (\ref{simpdefi}). Then we can similarly show that the derivative (\ref{gradformula}) is equivalent to the following expression:
\begin{align}
    \nabla c(\bm{S})&=-\frac{2 \gamma}{n} \sum_{i=1}^n  \bm{B}^T\bm{\gamma}_i(\bm{S})\bm{\gamma}_i(\bm{S})^T\bm{V} \\&=-\frac{2 \gamma}{n} \sum_{i=1}^n  \bm{B}^T_{W_i}\bm{\gamma}_{W_i}(\bm{S})\bm{\gamma}_{W_i}(\bm{S})^T\bm{V}_{W_i}. \label{simpprojimputederiv1}
\end{align}
The matrix multiplication in the final expression (\ref{simpprojimputederiv1}) has computational complexity of $O(m_i(p+k)+pk)$ (every individual term has been previously calculated), so summing over $n$ terms the computational complexity is 
$$O\left(|\Omega|\left(p+k+\frac{pk}{m\alpha}\right)\right).$$
Thus, the computational complexity of the entire step is:
$$O\left(|\Omega| \left(p+k^2+\frac{pk+k^3}{m\alpha}\right)\right).$$

\subsubsection*{Computational Complexity of Step \ref{calc_final}}
Recall that Step \ref{calc_final} of projImpute does the following calculation for every $i \in \{1,\cdots,n\}$:
\begin{equation}
    \bm{a}_i \leftarrow \overline{\bm{a}}_i\bm{B}\bm{S}^{*}(\bm{B}\bm{S}^{*}\bm{W}_i\bm{B}\bm{S}^{*T}\bm{B}^T)^{-1}\bm{S}^{*T}\bm{B}^T,
\end{equation}
where $\bm{S}^*$ is the imputed $\bm{S}$. Now define $\bm{V}^*=\bm{B}\bm{S}^{*}$, and define $\bm{V}^*_{W_i}$ in the same fashion as defined in (\ref{simpdefi}). Then we have that:
\begin{align}
    \overline{\bm{a}}_i\bm{B}\bm{S}^{*}(\bm{B}\bm{S}^{*}\bm{W}_i\bm{B}\bm{S}^{*T}\bm{B}^T)^{-1}\bm{S}^{*T}\bm{B}^T&=\overline{\bm{a}}_i\bm{V}^{*}(\bm{V}^{*T}\bm{W}_i\bm{V}^{*})^{-1}\bm{V}^{*T}\nonumber\\&=\bm{a}_{W_i}\bm{V}^{*}(\bm{V}^{*T}_{W_i}\bm{V}^{*}_{W_i})^{-1}\bm{V}^{*T}_{W_i} .\label{simpcalcfinal}
\end{align}
The final expression (\ref{simpcalcfinal}) has a matrix multiplication complexity of $O(km_i+k^2m_i+k^3+mk)$, so summing over $n$ samples the total computational complexity of Step \ref{calc_final} is:
$$O\left(|\Omega|\left(k^2+\frac{k}{\alpha}+\frac{k^3}{m\alpha}\right)\right).$$
\endproof
Using these results we can easily arrive at the following corollary:
\begin{cor}
The projImpute algorithm terminates in $t_{max}$ number of steps. Furthermore, it has complexity $$O\left(|\Omega| \left(p+k^2+\frac{pk+k^3}{m\alpha}\right)\right).$$
\end{cor}
\proof
The computational complexity follows immediately from Proposition \ref{prop:complexproj}.
\endproof
Note that with $p\gg k$ sufficiently large, the computational complexity of Step \ref{calc_cost} dominates that of Step \ref{calc_final}. In particular, such step scales scales linearly in $m$, $n$, and $p$, which is undesirable when all of these are large, as it is commonly in real-world settings. In the next section, we  design an algorithm that reduces the computational complexity on Step \ref{calc_cost}. 
\section{fastImpute: An Adaptive Stochastic Projected Gradient Descent Algorithm for Matrix Completion}
\label{sec:mc_pgdside}
In the previous section, we introduced  the projImpute algorithm, which conducted full projected gradient updates for every step $t \in \{1,\cdots, t_{max}\}$ using all $n$ rows and $m$ columns. However, it is computationally prohibitively expensive, especially when $n,m,p$ is large.

In this section, we introduce fastImpute that uses random sampling of rows and columns to estimate the gradient update at step $t$. Specifically, at each step, we randomly select $n_0$ rows and \emph{for each row}, we randomly select $m_0$ columns to calculate the sampled derivative for that row. Let $[n_t]$ denote the random set of numbers from $\{1,\cdots,n\}$ of size $n_0$ at step $t$, and $[m_t^i]$ denote a set of numbers from $\{1,\cdots,m\}$ of size $m_0$ at step $t$ corresponding to the random sample for row $i$. Then the objective function evaluated  with rows $[n_t]$ and columns $[m_t^i]$ is:
\begin{align*}
	c_{t}(\bm{S})=\frac{1}{n_0m_0} \sum_{i \in [n_t]} \overline{\bm{a}}_{im_t^i} 
	\left(\bm{I}_{m_t^i}-\bm{V}_{m_t^i}\left(\frac{\bm{I}_k}{\gamma}+\bm{V}_{m_t^i}^T\bm{W}_{im_t^i}\bm{V}_{m_t^i}\right)^{-1}\bm{V}_{m_t^i}^T\right)\overline{\bm{a}}_{im_t^i}^T,
\end{align*}
where $\bm{V}_{m_t^i}$ is the submatrix of $\bm{V}$ formed with $[m_t^i]$ columns, and similarly with other terms. We use the notation $m_t^i$ to explicitly indicate that the columns are selected for each row $i \in [n_t]$ and independent across rows.  
Similarly, using Lemma \ref{lem:derivresult}, the derivative evaluated with rows $[n_t]$ and columns $[m_t^i]$ is:
\begin{align*}
	\nabla c_{t}(\bm{S})&=\frac{1}{n_0m_0} \sum_{i \in [n_t]} -2 \gamma \bm{B}_{m_t^i}^T\left(I_{m_t^i}-\bm{V}_{m_t^i}\left(\frac{\bm{I}_k}{\gamma}+\bm{V}_{m_t^i}^T\bm{W}_{im_t^i}\bm{V}_{m_t^i}\right)^{-1}\bm{V}_{m_t^i}^T\right)\overline{\bm{a}}_{im_t^i}^T\\&\times\overline{\bm{a}}_{im_t^i}\left(I_{m_t^i}-\bm{V}_{m_t^i}\left(\frac{\bm{I}_k}{\gamma}+\bm{V}_{m_t^i}^T\bm{W}_{im_t^i}\bm{V}_{m_t^i}\right)^{-1}\bm{V}_{m_t^i}^T\right)^T\bm{V}_{m_t^i}.
\end{align*}
\begin{algorithm}[ht!] 
	\begin{algorithmic}[1]
	
	\Procedure{fastImpute}{$\bm{A},\bm{B}$,$k$,$\theta$, $t_{max}$}\Comment{$\bm{A} \in \md{R}^{n\times m}$ the masked matrix, $\bm{B} \in \md{R}^{p\times m}$ the feature matrix, $k$ the desired rank, $\theta$ the step size, and $t_{max}$ the number of gradient steps}
		\State $t \gets 1$
		\State $\alpha \gets \frac{|\Omega|}{mn}$ \Comment{Define existing percentage of $\bm{A}$. $|\Omega|$ is the set of non-zero entries in $\bm{A}$}
		\State $\bm{S}_1\gets \text{random initial matrix with $\|\bm{S}_1\|_2=1$}$ \Comment{Randomized Start}
		\State $\eta_0 \gets \infty $ \Comment{Initialize objective value}
		\State $q_1 \gets 0 $ \Comment{Initialize counter for non-improving steps}
		\State $m_0 \gets \min(2p,m)$ \Comment{Define initial gradient update size for columns}
		\State $n_0 \gets \left\lfloor\frac{k\sqrt{nm}\log (\sqrt{nm})}{8m_0\alpha}\right\rfloor$ \Comment{Define initial gradient update size for rows}
		\State $[n_0] \subset \{1,\cdots,n\} $ \Comment{Initialize rows selected}
		\For{$i \in [n_0]$}
		\State $[m_0^i] \subset \{1,\cdots,m\} $ \Comment{Initialize columns selected}
		\EndFor 
		\State $\eta_1, \bm{G}_1\gets c_{t}(\bm{S}_1), \nabla c_{t}(\bm{S}_1) $ \Comment{Initialize objective value and gradient}		
		\State $\nabla \tilde{\bm{S}}_{1} \gets \bm{0}^{k \times p}$ \Comment{Initialize accelerated gradient}
		\While{$t<t_{max}$}\Comment{While we have not reached $t_{max}$ iterations}
		\State $[n_t] \subset \{1,\cdots,n\} $ \Comment{Select new rows}
		\For{$i \in [n_t]$}
		\State $[m_t^i] \subset \{1,\cdots,m\} $ \Comment{Select new columns}
		\EndFor 
		\State $\nabla \tilde{\bm{S}}_{t+1}=\bm{G}_t + \frac{t-1}{t+2}\nabla \tilde{\bm{S}}_{t}$ \Comment{Nesterov accelerated gradient update step} 
		\State $\overline{\nabla \bm{S}_{t+1}} = -\nabla \tilde{\bm{S}}_{t+1} + (\nabla \tilde{\bm{S}}_{t+1} \cdot \bm{S}_t)\bm{S}_t$ \Comment{Project gradient to the tangent plane of $\bm{S}_t$} \label{step:project}
		\State $\bm{S}_{t+1} \gets  \bm{S}_t \cos \theta + \frac{\overline{\nabla \bm{S}_{t+1}}}{\|\overline{\nabla \bm{S}_{t+1}}\|_2} \sin \theta$ \Comment{Update $\bm{S}_t$ based on projected gradient}  \label{step:update}
		\State $\eta_{t+1}, \bm{G}_{t+1}\gets c_{t}(\bm{S}_{t+1}), \nabla c_{t}(\bm{S}_{t+1}) $ \Comment{Update the cost and derivative.}
		\State $t \gets t+1$
		\EndWhile
		\State $\bm{S}^*\gets \bm{S}_t$
		\State $i \gets 1$
		\For{$i<n$}
		\State $\bm{a}_i \gets \overline{\bm{a}}_i\bm{B}\bm{S}^{*}(\bm{B}\bm{S}^{*}\bm{W}_i\bm{B}\bm{S}^{*T}\bm{B}^T)^{-1}\bm{S}^{*T}\bm{B}^T$ \Comment{Calculate the final $\bm{A}$ matrix} 
		\EndFor
		\State \textbf{return} $\bm{A}$\Comment{Return the filled matrix $\bm{A}$}
		\EndProcedure
	\end{algorithmic}
	\caption{Gradient Descent algorithm for  matrix completion with side information. }
	\label{alg:mc_alg2}
\end{algorithm}

We then present the algorithm as Algorithm \ref{alg:mc_alg2}.
It is the same procedure as projImpute, except with stochastic gradient updates of $n_0$ rows and $m_0$ columns. 
For this algorithm to work, we  need the stochastic gradient to be ``close'' to the true gradient, which is true using a theorem adapted from \cite{bertsimas2018interpretable}:
\begin{theo}
\label{theo:gradapprox}
    Let $\bm{A}$ be a partially known matrix, $\bm{B}$ a known feature matrix, and $\bm{W}_i$ as defined in Proposition \ref{prop:maintheo}. Then, with probability at least $1-\epsilon$, for any $t$, and any $\bm{S}$ satisfying $\|\bm{S}\|=1$, we have:
    \begin{align*}
        |c_t(\bm{S})-c(\bm{S})|&\leq \sqrt{\frac{Ak\log\left(\frac{k}{\epsilon}\right)}{n_0}} ,\\
                \|\nabla c_t(\bm{S})-\nabla c(\bm{S})\|_2&\leq \sqrt{\frac{Bk\log\left(\frac{k}{\epsilon}\right)}{n_0}}.
    \end{align*}
\end{theo}
Inverting the statements, we have that:
\begin{align*}
    \P(|c_t(\bm{S})-c(\bm{S})|>\epsilon)&\leq ke^{-\frac{n_0\epsilon^2}{Ak}}, \vspace{3pt}\\
   \P(\|\nabla c_t(\bm{S})-\nabla c(\bm{S})\|>\epsilon)&\leq ke^{-\frac{n_0\epsilon^2}{Bk}}    .
\end{align*}
Theorem \ref{theo:gradapprox}   shows that the probability the stochastic objective value and its derivative  is $\epsilon$ away from the true value is exponentially vanishing with increasing $n_0$. 
This  property   affects how we choose $n_0$ and $m_0$, which is discussed in Section \ref{subsec:fastimputescomp}. 

In the next  section, we   show that  fastImpute 
 recovers the true solution for matrix completion under relatively mild conditions. 
\subsection{Convergence Rates and Guarantees}
\label{subsec:convergence}
In this section, we  derive results that show  Algorithm \ref{alg:mc_alg2} converges to the global minimum 
of $c(\bm{S})$. for the case  $1/\gamma=0$ and  without side information ($p=m$ and $\bm{B}=\bm{I}_m$). Under such assumptions,  our objective can be written as:
\[c(\bm{S})=\frac{1}{nm}\sum_{i=1}^n  \overline{\bm{a}}_i
\left(\bm{I}_m-\bm{S}\left(\bm{S}^T\bm{W}_i\bm{S}\right)^{-1}\bm{S}^T\right)\overline{\bm{a}}_i^T.\]
The analysis  can be easily extended to the case where $\bm{B}$ is a general feature matrix, at the expense of increased notational complexity. The analysis below also holds for $\frac{1}{\gamma}<\epsilon_0$ for some sufficiently small $\epsilon_0$ with minimal changes. To facilitate the theoretical analysis, we would further assume that the final projected gradient is perturbed by some asymptotically vanishing Gaussian noise (immediately before Step \ref{step:update}):
\[\overline{\nabla \bm{S}_{t+1}} \gets \overline{\nabla \bm{S}_{t+1}} + O\left(\frac{1}{\sqrt{n_0m}}\right)\bm{E}\]
Where $\bm{E}\in \md{R}^{k \times p}$ has iid entries with the standard normal distribution. As $n_0$ and $m$ is large, this does not significantly change the resultant gradient, and numerical experiments suggest that this has no appreciable impact on the final solution. 

Over the last few years, various researchers have proven related results using different formulations for matrix completion (see  \cite{ge2016matrix,ma2019implicit} for the case where the recovered matrix is semi-definite, and \cite{zheng2016convergence} for the general case using a lifting formulation). The proofs  follow a two-step process:
\begin{itemize}
	\item Prove the proposed algorithm converges to a local minimum of the objective function.
	\item Prove all local minima of the objective function are global minima (equivalently, the objective function has ``no spurious local minima"). This condition has been shown to be true for various formulations of nonconvex low rank problems, as explored in e.g. \cite{ge2017no}. 
\end{itemize}
In our new formulation of matrix completion, we  show how proving the first result is equivalent to checking the ``strict saddle" condition on the objective function introduced in \cite{ge2015escaping}. We  then prove the objective function $c(\bm{S})$ does indeed satisfy the strict saddle condition, and moreover has no spurious local minima. 

Before we layout our detailed results, we need to specify a random sampling model for the known entries of $\bm{A}$. Generally in the literature (see e.g. \cite{candesexact,jin2016provable,ge2016matrix} for examples),
 the uniform sampling model is used, where every element is assumed to be present independently with a probability $r_0$.  To simplify the proof,  we  specify a row-sampling model: For every row $\bm{a}_i$ of $\bm{A}$, we randomly sample (without replacement) $l$ out of $m$ entries to be known. Thus, each entry has a probability of $r=\frac{l}{m}$ being selected. 
   By results in \cite{candesexact} and \cite{gross2010note}, it is known that convergence results under row-sampling models with sampling rate $r$ hold true under the uniform sampling model with a rate $r_0=O(\log n)r$, where the factor $\log n$ is essentially due to the coupon collector effect (for more discussion please see \cite{candesexact}). 

We next  show that proving Algorithm  \ref{alg:mc_alg2} converges to a local minimum is equivalent to verifying $c(\bm{S})$ satisfies the ``strict saddle" condition. We first formally define such concept, as introduced in \cite{ge2015escaping}:
\begin{defi}
	A function $f: \md{R}^d \to \md{R}$  is $(\theta, \zeta, \eta)$-strict saddle if for every $\bm{x}$, at least one of the following hold for constants $\theta, \zeta, \eta>0$:
	\begin{enumerate}
		\item $\|\nabla f(\bm{x})\|\geq \theta$.
		\item $\lambda_{\text{min}}(\nabla^2f(\bm{x}))\leq -\zeta$.
		\item There exists a local minimum $\bm{x}^*$ such that $\|\bm{x}-\bm{x}^*\|\leq \eta$. 
	\end{enumerate}
\end{defi} 
Using this definition, the following theorem from  \cite{fang2019sharp} (see also \cite{daneshmand2018escaping,ge2015escaping}) establishes that a stochastic gradient algorithm converges to a local minimum if the function $f$ being optimized over satisfies the strict saddle condition:
\begin{theo}
	\label{theo:gdconverge}
	Assume that  $f(\bm{x})$ is a function of $\bm{x} \in \md{R}^d$ that is a $(\theta, \zeta, \eta)$-strict saddle. Then, a stochastic gradient descent algorithm  with stochastic gradient $\tilde{\nabla} f(\bm{x})$ and added Gaussian noise $\frac{\sigma}{\sqrt{d}}\bm{\epsilon}$ where $\bm{\epsilon} \sim N(0,\bm{I}_d)$ converges in $O\left(\poly\left(\frac{1}{\delta}\right)\right)$ iterations to a point $\delta$ close to a local minimum, where we have:
    \begin{equation*}
\P(\|\tilde{\nabla} f(\bm{x})-\nabla f(\bm{x})\|>\epsilon)\leq e^{-\frac{\epsilon^2}{2\sigma^2}}   .
\end{equation*}\end{theo}
In the current problem,  Theorem \ref{theo:gradapprox} shows that a valid bound on the standard deviation is $\sqrt{\frac{Bk\log(k)}{n_0}}$. Therefore, as $d=mk$, a stochastic gradient descent algorithm with noise of $O(\frac{1}{\sqrt{n_0m}})$ would converge to a local minimum of $c(\bm{S})$. However,  Algorithm \ref{alg:mc_alg2} is a \emph{projected} stochastic gradient descent algorithm (with such added perturbation), so we cannot immediately apply Theorem \ref{theo:gdconverge}. Thus, we next  connect Algorithm \ref{alg:mc_alg2} with the standard stochastic gradient algorithm. Note that for the case $1/\gamma=0$, the stochastic gradient is always orthogonal to $\bm{S}$:
\begin{lem}
	Let $1/\gamma=0$. Then for all $\bm{S}$ and any random sample $[n_t],[m_t]$ of rows and columns, we have
	\[\bm{S}^T\nabla c_t(\bm{S})=0.\]
\end{lem}
The statement follows immediately from Lemma \ref{lem:first_deriv} in Appendix \ref{app:localminproof}. Therefore, the projection step (Step \ref{step:project} in Algorithm \ref{alg:mc_alg2}) is trivial and reduces to:
\[\overline{\nabla \bm{S}_{t+1}} = -\nabla \tilde{\bm{S}}_{t+1} \]
In other words, the projected gradient is exactly the original gradient, and the projection step does not change the derivative. Thus, Algorithm \ref{alg:mc_alg2} under $1/\gamma=0$ reduces to a stochastic gradient algorithm. 

Thus, if we can establish that $c(\bm{S})$ is a $(\theta,\zeta, \eta)$-strict saddle, we can indeed use Theorem \ref{theo:gdconverge} to prove that Algorithm \ref{alg:mc_alg2} converges to a local minimum. Therefore, to establish the global convergence of Algorithm \ref{alg:mc_alg2}, we now equivalently need to establish the following two results:
\begin{enumerate}
    \item The function $c(\bm{S})$ is a $(\theta, \zeta, \eta)$-strict saddle. 
    \item All local minima of $c(\bm{S})$ are global minima.
\end{enumerate}
These properties and Theorem  \ref{theo:gdconverge} will then establish that 
Algorithm \ref{alg:mc_alg2}  converges to a point $\delta$-close to the global minimum in $O\left(\poly\left(\frac{1}{\delta}\right)\right)$ iterations. We prove these two results simultaneously. To do so, we introduce a few regularity conditions that are necessary for these statements to be true. First, the matrix $\bm{A}$ we want to recover cannot be the following:
\begin{equation}
\bm{A}=\begin{pmatrix}
1 & 0 & \cdots & 0\\
0 & 0 & \cdots & 0\\
\vdots & \vdots & \ddots & \vdots \\
0 & 0 & \cdots & 0
\end{pmatrix}. \label{eq:badmatrix}
\end{equation}
This is because to recover such matrix truthfully, we  need to have to know $A_{11}$ as knowing any other element would give us no information about the 1 in the top left corner. To know $A_{11}$ we  
would require basically all entries  to be known under a random sampling model, which is undesirable; see \cite{candes2009exact} for further discussion. To prevent this from happening, we introduce the following concept:
\begin{defi}
We define a matrix $\bm{S} \in \md{R}^{m \times k}$, for $m>k$, to be $(\alpha,p)$-submatrix full rank
if every $p \times k$ submatrix of $\bm{S}$ is full column rank and the minimum singular value is at least $\alpha $. 
\end{defi} 
We introduce the following assumption:
\begin{assumption}
\label{ass:ripweak}
The desired matrix to be recovered, $\bm{A}$, is of rank $k$ and admits a decomposition $\bm{A}=\bm{U}^*\bm{S}^{*T}$ where $\bm{U}^*\in \md{R}^{n \times k}, \bm{S}^* \in \md{R}^{m \times k}$, $\|\bm{S}^*\|=1$, and $\bm{U}^*$, $\bm{S}^*$ is $(\alpha, l)$-submatrix full rank, where $l/m$ is the sampling rate and  $\alpha>0$. 
\end{assumption}
Qualitatively, this assumption requires the decomposition of $\bm{A}$ to be ``spread out" enough so that no particular row and column is special. In particular, it excludes the pathological matrices like the one presented in Eq. \eqref{eq:badmatrix}, as $\bm{A}=(1,0,\cdots,0)(1,0,\cdots,0)^T=\bm{U}^*\bm{S}^{*T}$, and most of the submatrices of $\bm{U}^*=\bm{S}^*=(1,0,\cdots,0)$ have rank 0, save for those that includes the first element. This is because the first element is special in the decomposition of such $\bm{A}$, and if the first element is missing, no information for $\bm{A}$ can be deduced.  Assumption \ref{ass:ripweak} excludes these matrices so that no element is special enough such that if it is missing, we fail to recover $\bm{A}$. 

We now introduce a second technical assumption. If $\bm{W_i}\bm{S}$ has rank $<k$, then the inverse $(\bm{S}^T\bm{W}_i\bm{S})^{-1}$ in the objective function $c(\bm{S})$ would not be well-defined, so the objective function is not defined. Therefore, the second assumption is that we restrict our domain to the set of $\bm{S}$ where the objective function is defined:
\begin{assumption}
\label{ass:Vrestrict}
The set $\bm{S}$ is restricted to matrices that are  $(\alpha, l)$-submatrix full rank, where $l/m$  is the sampling rate and  $\alpha>0$. 
\end{assumption}
In real-world situations this assumption is almost never violated as numerical matrices almost never have \emph{perfectly} singular submatrices. If it is, any small random perturbation around $\bm{S}$ would almost surely make the condition true. 

Now, with these regularity conditions, we establish that for $r$ sufficiently large, the two results we aim to prove are true:
\begin{theo}
\label{theo:localmin}
 Assume $\bm{A}$ follows the row-sampling model, and each element is known with probability $r=\frac{l}{m}>\frac{Ck}{m}$ for sufficiently large $C$. Further assume Assumptions \ref{ass:ripweak}-\ref{ass:Vrestrict} hold. Then, with probability $1-O(\frac{1}{n})$, for $1/\gamma=0$ and $\bm{B}=\bm{I}_m$, $c(\bm{S})$ is $(\epsilon, O(\frac{\epsilon}{k}),O(\sqrt{\epsilon}))$-strict saddle for sufficiently small $\epsilon$. Furthermore, all local minima of $c(\bm{S})$ are global minima. 
\end{theo}
The proof is contained in Appendix \ref{app:localminproof}. Then combining Theorem \ref{theo:localmin} with Theorem \ref{theo:gdconverge}, we have the desired global convergence result:
\begin{cor}
Consider the minimization problem defined in (\ref{eq:mc_central}) under $1/\gamma=0$ and $\bm{B}=\bm{I}_m$. Assume that Assumptions \ref{ass:ripweak}-\ref{ass:Vrestrict} hold. Then, under the row-sampling model, for $r>\frac{Ck}{m}$ with $C$ sufficiently large, Algorithm \ref{alg:mc_alg2} (with added perturbation) outputs a point that is $\delta$-close to a global minimum in $\poly(\frac{1}{\delta})$ iterations with probability $1-O(\frac{1}{n})$. 
\end{cor}
As explained above, we can transfer results in the row-sampling model to the commonly utilized uniform sampling (Bernoulli) model:
\begin{cor}
Consider the minimization problem defined in (\ref{eq:mc_central}) under $1/\gamma=0$ and $\bm{B}=\bm{I}_m$. Assume that Assumptions \ref{ass:ripweak}-\ref{ass:Vrestrict} hold. Then, under the uniform sampling model, for a sampling rate $r_0 >\frac{Cnk\log(n)}{mn}$ with $C$ sufficiently large, Algorithm \ref{alg:mc_alg2} (with added perturbation) outputs a point that is $\delta$-close to a global minimum in $\poly( \frac{1}{\delta})$ iterations with probability $1-O(\frac{1}{n})$. 
\end{cor}
The conclusion is consistent with other algorithms which require $O(nk^\alpha \log(n))$ samples under the uniform sampling model. 
\subsubsection{Necessity of Regularization}
\label{subsubsec:regularization}

The previous results deal with the case when there is no regularization ($1/\gamma=0$) though it can be readily extended to any sufficiently small regularization term. Therefore, it is natural to question the purpose of the $\ell_2$ regularization term in the formulation. The answer is two-fold:
\begin{enumerate}
    \item The proof above is for the case when the matrix $\bm{A}$ is perfectly low rank.  However, in real-world cases, the data might be corrupted in unseen ways, and it is shown in \cite{bertsimas2018characterization} that $\ell_2$ regularization provides robustness properties that could guard against such corruption. 
    \item There are cases in the real-world in which the the number of known entries in a row $\bm{A}_i$ of the partially-known matrix is less than the desired rank $k$, and in such cases the regularization becomes necessary to recover the true solution, as $\rank(\bm{W}_i)<k$, so $\bm{S}\bm{W}_i\bm{S}^T$ is not invertible. Adding a regularization term guards against such cases. 
\end{enumerate}
\subsection{Computational Complexity of fastImpute}
\label{subsec:fastimputescomp}
Here we  first discuss how $m_0$ and $n_0$ is chosen.
By Theorem \ref{theo:gradapprox} on the approximation of the stochastic gradient and objective, the error in such approximation drops exponentially with increasing $n_0$. Therefore, we 
 want to ensure $n_0$ is never too small. 

Now, by our convergence result in Theorem \ref{theo:localmin}, we need $O(n k \log n)$ samples to recover the true solution under the uniform sampling model, when $n$ is sufficiently large. Since $\alpha$ is the empirical probability that any element is known, we need to have:
\[m_0n_0\alpha = O(n k \log n).\]
The complexity required to discern the $p$ factors we can choose from is $O(p)$, thus, our selection of $n_0$ and $m_0$ is:
$$m_0=O(p) , \qquad n_0=\max\left\{O\left(\frac{nk\log (n)}{m_0\alpha}\right),\underline{n}\right\}.$$
where $\underline{n}$ is some lower bound on $n_0$ to ensure that Theorem \ref{theo:localmin} can be applied. We now show that such $m_0$ and $n_0$ reduces the computational complexity of the gradient update step asymptotically:
\begin{cor}
\label{cor:fastimputecomp}
The computational complexity of Step \ref{calc_cost} in Algorithm \ref{alg:mc_alg2} is $$O\left(\frac{(pk+k^3)n\log (n)}{\alpha}\right).$$
\end{cor}
\proof
For the gradient update step, we replace $m$ with $O(p)$ and $n$ with $O\left(\frac{kn\log (n)}{p\alpha}\right)$ in the original formula for Step \ref{calc_cost} in projImpute to get that its computational complexity is:
\begin{align*}
    &O\left(\frac{kn\log (n)}{\alpha}\left(p+k^2+\frac{k}{\alpha}+\frac{k^3+pk}{p\alpha}\right)\right)
    \\&=O\left(\frac{(pk+k^3)n\log (n)}{\alpha}\right).
\end{align*}
Where we have suppressed the last two terms as they are dominated by the first two.
\endproof
We note that the dependence of $m$ is completely removed (for sufficiently large $n$) when compared with Algorithm \ref{alg:mc_alg1}. This, however, does not reduce the asymptotic complexity with regards to $m$ for the full algorithm as eventually the step to fill the matrix $\bm{A}$ dominates, and that scales linearly with $m$. Our dependence on $k$ is cubic, but $k$ is usually small in real-world applications, so it is not a large concern. 

In the next section, we discuss experiments   for fastImpute.
\section{Experiments on fastImpute with Side Information}
\label{sec:mc_computeside}
In this section, we  compare fastImpute with other inductive matrix completion algorithms on both synthetic datasets and real-world datasets to explore its performance and scaling behavior.
\subsection{Synthetic Data Experiments}
 For synthetic data experiments, we assume that the underlying matrix satisfies the form $\bm{A}=\bm{U}\bm{S}^T\bm{B}^T$, where $\bm{U} \in \md{R}^{n \times k}$, $\bm{S} \in \md{R}^{p\times k}$, and $\bm{B} \in \md{R}^{m\times p}$. The elements of $\bm{U}$, $\bm{S}$, and $\bm{B}$ are selected from a uniform distribution of $[0,1]$, where a fraction  $\mu$ is missing. We report statistics on various combinations of $(m,n,p,k,\mu)$.

All algorithms are tested on a server with $16$ CPU cores. For each combination $(m,n,p,k,\mu)$, we ran 10 tests and report  the average value for every statistic.
The algorithms tested are:
 \begin{itemize}
     \item \textbf{fastImpute}: We use the sampling parameters:
     $$m_0=\min(2p,m), \qquad n_0=\max\left\{\frac{nk\log (n)}{8m_0\alpha},100\right\}.$$
     with $t_{max}=50$, and $\theta=\frac{\pi}{64}$, and regularization parameter $\gamma=10^6$. We explicitly stress here that no parameter tuning is done on fastImpute as we intend to show that the algorithm is not parameter sensitive.   We implement our algorithm in Julia 0.6 with only the base packages.
     \item \textbf{IMC}: This algorithm is a well-accepted benchmark for testing Inductive Matrix Completion algorithms developed by \cite{natarajan2014inductive}. For each combination of $(m,n,p,k,\mu)$, we tune the regularization parameter $\lambda$ by imputing random matrices with such combination and find the $\lambda$ that gives the best results. We utilize the implementation provided by the authors in Matlab.
 \end{itemize}
To further understand the benefits of the stochastic algorithm, we also compare against projImpute, the full gradient descent algorithm. We report the following statistics for each algorithm:
\begin{itemize}
	\item $n,m$ - the dimensions of $\bm{A}$.
	\item $p$ - the number of features in the feature matrix.
	\item $k$ - the true number of features.
	\item $\mu$ - The fraction of missing entries in $\bm{A}$.
	\item $T$ - the total time of algorithm execution.
	\item MAPE - the Mean Absolute Percentage Error (MAPE) for the retrieved matrix $\hat{\bm{A}}$:
	\[\text{MAPE}=\frac{1}{nm} \sum_{i=1}^n \sum_{j=1}^m \frac{|\hat{A}_{ij}-A_{ij}|}{|A_{ij}|}.\]
\end{itemize}

\begin{table}[ht!]
\centering
\begin{tabular}{|c|l|l|l|l|l||l|c|l|c|l|c|}
	\hline \multirow{2}{*}{} & \multirow{2}{*}{$\bm{n}$}&\multirow{2}{*}{$\bm{m}$}&\multirow{2}{*}{$\bm{p}$}&\multirow{2}{*}{$\bm{k}$} & \multirow{2}{*}{$\bm{\mu\%}$} & \multicolumn{2}{c|}{\textbf{projImpute}} & \multicolumn{2}{c|}{\textbf{fastImpute}} & \multicolumn{2}{c|}{\textbf{IMC}}\\\cline{7-12}
	&  &  & & & & $\bm{T}$ & \textbf{MAPE} & $\bm{T}$ & \textbf{MAPE}& $\bm{T}$ & \textbf{MAPE}\\\hline
	\multirow{4}{*}{$n$}&$10^3$ & $10^3$ & 100& 5 & $95\%$ & 5.1s &$0.2\%$& 3.1s &$0.4\%$ &4.4s & $2.9\%$ \\
	&$10^4$ & $10^3$& 100& 5 & $95\%$ & 56.7s&$0.1\%$& 12.0s&$0.1\%$ & 18s&$2.5\%$ \\
	&$10^5$ & $10^3$& 100 & 5 & $95\%$ & 580s &$0.2\%$& 54.8s &$0.2\%$ &173s & $1.0\%$\\
	&$10^6$ & $10^3$& 100 & 5 & $95\%$& 6035s &$0.2\%$ & 390s &$0.2\%$ &1400s & $0.3\%$  \\\hline\hline
	\multirow{4}{*}{$m$}	&$10^4$ & $10^3$& 100& 5 & $95\%$ & 56.7s&$0.1\%$& 12.0s&$0.1\%$ & 18s&$2.5\%$ \\

	&$10^4$ & $10^4$& 100& 5 & $95\%$ & 605s&$0.2\%$& 39s&$0.4\%$ & 144s &$0.8\%$ \\
	&$10^4$ & $10^5$& 100 & 5 & $95\%$ & 5874s &$0.1\%$& 360s &$0.2\%$ & 1265s & $0.4\%$  \\
	&$10^4$ & $10^6$& 100 & 5 & $95\%$ & N/A &$0.1\%$& 3056s &$0.1\%$ & 13891s & $0.2\%$  \\\hline\hline
	\multirow{4}{*}{$p$}	&$10^4$ & $10^3$& 100& 5 & $95\%$ & 56.7s&$0.1\%$& 12.0s&$0.1\%$ & 18s&$2.5\%$ \\

	&$10^4$ & $10^3$& 200 & 5 & $95\%$ & 104s&$0.05\%$ & 23s&$0.1\%$ & 30s &$2.7\%$ \\
	&$10^4$ & $10^3$& 500 & 5 & $95\%$ & 230s &$0.06\%$& 40s &$0.05\%$ & 87s & $2.1\%$  \\
	&$10^4$ & $10^3$& 1000 & 5 & $95\%$ & 417s &$0.02\%$ & 85s &$0.01\%$ & 160s & $1.9\%$  \\\hline\hline
	\multirow{4}{*}{$k$}	&$10^4$ & $10^3$& 100& 5 & $95\%$ & 56.7s&$0.1\%$& 12.0s&$0.1\%$ & 18s&$2.5\%$ \\
	&	$10^4$ & $10^3$& 100 & 10 & $95\%$ & 194s&$0.1\%$& 37.6s&$0.2\%$ &32s & $2.1\%$\\
&	$10^4$ & $10^3$ & 100& 20 & $95\%$ & 460s&$0.2\%$& 90s&$0.4\%$ &47s & $2.2\%$\\
	&	$10^4$ & $10^3$& 100  & 30 & $95\%$ & 1055s &$0.3\%$ & 290s &$0.5\%$ &55s & $2.6\%$\\\hline\hline
\multirow{4}{*}{$\mu$} &	$10^4$ & $10^3$& 100 & 5 & $20\%$  & 107.4s &$0.5\%$& 25.1s &$0.7\%$ & 6.2s & $0.01\%$ \\
	&$10^4$ & $10^3$& 100 & 5 & $50\%$ & 92.6s &$0.2\%$& 19.2s &$0.4\%$ &8.9s & $0.03\%$ \\
	&$10^4$ & $10^3$& 100 & 5 & $80\%$ & 70.4s&$0.1\%$ & 15.7s&$0.1\%$ &14s & $0.7\%$\\
	&$10^4$ & $10^3$& 100& 5 & $95\%$ & 56.7s&$0.1\%$ & 12.0s&$0.1\%$ & 18s&$2.5\%$ \\\hline\hline
	&$10^4$ & $10^3$& 100& 5 & $95\%$ & 56.7s&$0.1\%$& 12.0s&$0.1\%$ & 18s&$2.5\%$ \\
	&$10^5$ & $10^3$& 100& 10 & $95\%$ & 1095s &$0.1\%$& 84s &$0.1\%$ & 203s & $0.8\%$\\
	&$10^5$ & $10^4$& 200& 10 & $95\%$ & 14506s &$0.1\%$ & 560s &$0.2\%$ &$2430s$ &$0.4\%$\\
	&$10^5$ & $10^5$& 200& 10 & $95\%$ & N/A & N/A & 3170s &$0.2\%$ &$25707s$ & $0.2\%$ \\
	&$10^6$ & $10^4$& 500& 20 & $95\%$ & N/A & N/A & 6516s &$0.2\%$ & N/A & N/A \\\hline\hline
\end{tabular}
\caption{Comparison of fastImpute and IMC on synthetic data.  $N/A$ means the algorithm did not complete running in 20 hours, corresponding to 72000 seconds. }
\end{table}
We have the following observations:
\begin{itemize}
	\item fastImpute significantly improves the scaling behavior of projImpute in $n$ and $m$, and reduces the running time by over $10$ times on real-world datasets. 
    \item For small $n,m$, fastImpute scales roughly linearly in $n$ and is not sensitive towards $m$. Eventually it scales as $O(nm)$ when the matrix filling step dominates, as predicted. IMC scales similarly with slightly worse accuracy.
    \item Both algorithms roughly exhibit linear scaling in $p$, but IMC retrieves a matrix that has much higher MAPE than fastImpute.
    \item fastImpute roughly shows the $O(k^3)$ dominant behavior as expected, while IMC seems to scale linearly in $k$.
    \item IMC's running time increases with more missing data while it decreases for fastImpute. Both algorithms achieve roughly the same performance, with IMC dropping significantly as the number of missing entries increased.
\end{itemize}
\subsection{Real-World Experiments}
\label{sec:fastImputeSrealexp}
For real-world experiments, we utilize the Netflix Prize Dataset. This dataset was released in a competition to predict ratings of customers on unseen movies, given over 10 million ratings scattered across $500,000$ people and $16,000$ movies. Thus, when presented in a matrix $\bm{A}$ where $A_{ij}$ represents the rating of individual $i$ on movie $j$, the goal is to complete the matrix $\bm{A}$ under a low-rank assumption. 

For this experiment, we  included movies where people who had at least $5$ ratings present. This gives a matrix of  $471,268$ people and $14,538$ movies. To observe the scalability of fastImpute, we created five data sets (in similar format to \cite{bertsimas2018interpretable}):
\begin{enumerate}
	\item Base - $\bm{A}_1$ has  dimensions   $3,923\times 103$.
	\item Small - $\bm{A}_2$ has  dimensions   $18,227\times 323$.
	\item Medium - $\bm{A}_3$ has  dimensions   $96,601\times 788$.
	\item Large - $\bm{A}_4$ has  dimensions   $471,268 \times 1760$.
	\item Full - $\bm{A}$ has  dimensions   $471,268\times 14,538$. 
\end{enumerate}
These sizes are constructed such that the total number of elements in $\bm{A}$ in the successive sizes are approximately different by approximately an order of magnitude.

The feature matrix $\bm{B}$  is constructed using data from the TMDB Database, and covers 59 features that measure geography, popularity, top actors/actresses, box office, runtime, genre and more. The full list of 59 features is contained in  Appendix \ref{app:featurelist}.

For comparison, we test against IMC. We split the training set in $80\%/20\%$, where the latter group is used for validation of the rank in IMC and fastImpute. We then report the time taken, $T$, the MAPE, and the optimal chosen rank $k^*$ for each algorithm:
\begin{table}[h!t]
	\centering
	\resizebox{\columnwidth}{!}{%
	\begin{tabular}{|l|l|l|l||l|l|l|l|l|l|}
		\hline  \multirow{2}{*}{$\bm{n}$}&\multirow{2}{*}{$\bm{m}$} & \multirow{2}{*}{$\bm{p}$} & \multirow{2}{*}{$\bm{\mu\%}$} &\multicolumn{3}{c|}{\textbf{fastImpute}} & \multicolumn{3}{c|}{\textbf{IMC}}\\\cline{5-10}
		 &  &  &  & $\bm{T}$  & $\bm{k^*}$ & \textbf{MAPE}  & $\bm{T}$  & $\bm{k^*}$ & \textbf{MAPE} \\\hline
			3,923 & 103  & 59 & $92.6\%$ & 1.7s & 5 & $32.9\%$ &  0.8s & 5 & $34.1\%$ \\
		18,227 & 323  & 59 & $94.8\%$ & 11 & 6 & $28.0\%$ & 7.5s & 6 & $29.0\%$\\
		96,601 & 788  & 59& $94.2\%$ & 75s & 7 & $25.7\%$ & 49s & 8 & $28.5\%$\\
		471,268 & 1,760 & 59 & $93.6\%$ & 460s & 8 & $22.9\%$ & 870s & 10 & $24.1\%$ \\
		471,268 & 14,538 & 59 & $94.1\%$ & 2934s & 8 & $20.7\%$ & 7605s & 10 & $21.0\%$\\\hline
	\end{tabular}
	}
	\caption{Comparison of methods on Netflix data for fastImpute.}
\end{table}
We see that fastImpute is able to outperform IMC on the Netflix dataset across the different $n$ and $m$ values, while enjoying competitive scalability.

\section{fastImpute without Side Information}
\label{sec:mc_computegeneral}
In this section, we  explore fastImpute in the special case where there is no side information ($p=m$, and $\bm{B}=\bm{I}_m$). In such case, the objective function becomes
\[c(\bm{S})=\frac{1}{nm}\sum_{i=1}^n  \overline{\bm{a}}_i
	\left(\bm{I}_m-\bm{S}\left(\bm{S}^T\bm{W}_i\bm{S}\right)^{-1}\bm{S}^T\right)\overline{\bm{a}}_i^T.\]
This special cases give rises to two additional optimizations:
\begin{itemize}
    \item Define $d=\min\{m,n\}$. Since the problem is now symmetric in $n$ and $m$, instead of requiring $O(nk\log(n))$ samples, we only require $O(dk\log(d))$ as we can similarly perform the matrix completion on $\bm{A}^T$. 
    \item The multiplications involving $\bm{B}$ are no longer needed as $\bm{B}$ is the identity matrix and thus does not contribute. 
\end{itemize}
With these two observations, we can update our corollary  about the computational complexity of the gradient update step:
\begin{cor}
For $m=p$, and $\bm{B}=\bm{I}_m$, the computational complexity of Step \ref{calc_cost} in Algorithm \ref{alg:mc_alg2} is $$O\left(\frac{k^3d\log (d)}{\alpha}\right).$$
\end{cor}
Using this further optimization, we compare fastImpute with multiple general matrix completion algorithms on both synthetic datasets and real-world datasets to explore its performance and scaling behavior.
\subsection{Synthetic Data Experiments}
 For synthetic data experiments, we assume that the underlying matrix satisfies the form $\bm{A}=\bm{U}\bm{S}^T$, where $\bm{U} \in \md{R}^{n \times k}$, $\bm{S} \in \md{R}^{m\times k}$. Then the elements of $\bm{U}$ and $\bm{S}$ are selected from a uniform distribution of $[0,1]$, where a fraction  $\mu$ is missing. We report statistics on various combinations of $(m,n,k,\mu)$.
 
 The algorithms tested are:
 \begin{itemize}
     \item \textbf{fastImpute}: We use the sampling parameters:
     $$m_0=m, \qquad n_0=\max\left\{\frac{nk\log (n)}{4m_0\alpha},100\right\}.$$
     with $t_{max}=50$, and $\theta=\frac{\pi}{64}$, and regularization parameter $\gamma=10^6$. We explicitly stress here that no parameter tuning is done on fastImpute as we intend to show that the algorithm is not parameter sensitive. We implement our algorithm in Julia 0.6 with only the base packages.
     \item \textbf{softImpute-ALS (SIALS)}: Developed by \cite{softimputeals}, this is widely recognized as a state-of-the-art matrix completion method without feature information. It has among the best scaling behavior across all classes of matrix completion algorithms as it utilizes fast alternating least squares to achieve  scalability. For each combination of $(m,n,\mu,k)$, we tune the regularization parameter $\lambda$ by imputing random matrices with such combination and find the $\lambda$ that gives the best results. We utilize the implementation in the softImpute package in R for testing.
     \item \textbf{softImpute-SVD (SISVD)}: Developed by \cite{SoftImpute}, this is the original softImpute algorithm that utilizes truncated SVDs and spectral regularization to impute the matrix. This method is used as a fast benchmark for SVD-type methods. For each combination of $(m,n,\mu,k)$, we tune the regularization parameter $\lambda$ by imputing random matrices with such combination and find the $\lambda$ that gives the best results. We utilize the implementation in the softImpute package in R for testing.
     \item \textbf{Matrix Factorization Stochastic Gradient Descent (MFSGD)}: This is a popular stochastic gradient descent algorithm (discussed in \cite{jin2016provable}) which separates $\bm{A}=\bm{U}\bm{V}$ , where $\bm{U} \in \md{R}^{n \times k}$ and $\bm{V}\in \md{R}^{k \times m}$), and perform gradient updates for $\bm{U}$ and $\bm{V}$. We utilize the implementation in the Fancyimpute package of python that utilizes Tensorflow and the latest available speed optimizations for such algorithm. 
 \end{itemize}
To further illustrate the favorable scaling behavior of fastImpute, we 
 compare fastImpute against \emph{online} matrix completion algorithms. Online matrix completion algorithms complete the matrix by updating the factorization $\bm{U}$ and $\bm{V}$ with sequential data input from $\bm{A}$ over time. By their sequential nature, they are usually much faster than offline matrix completion algorithms which consider all the data  jointly.  The specific online algorithm we  compare to is:
\begin{itemize}
	\item \textbf{GROUSE}: This is a popular and efficient online matrix completion algorithm based on sequential gradient updates on the Grassmann manifold, as set out in \cite{balzano2010online}. We utilize the official implementation in MATLAB along with the latest optimizations in MATLAB R2020a. 
\end{itemize}
We 
 note that online matrix completion is usually used in different settings than offline algorithms and such comparison is only to note the favorable scaling behavior of fastImpute.

 All of the algorithms are executed on a server with $16$ CPU cores. Each combination $(m,n,k,\mu)$ was ran 10 times, and we report the average value of every statistic. The statistics reported are as followed:
\begin{itemize}
	\item $n,m$ - the dimensions of $\bm{A}$.
	\item $k$ - the true number of features.
	\item $\mu$ - The fraction of missing entries in $\bm{A}$.
	\item $T$ - the total time of algorithm execution.
	\item MAPE - the Mean Absolute Percentage Error (MAPE) for the retrieved matrix $\hat{\bm{A}}$:
	\[\text{MAPE}=\frac{1}{nm} \sum_{i=1}^n \sum_{j=1}^m \frac{|\hat{A}_{ij}-A_{ij} |}{|A_{ij}|}. \]
\end{itemize}
The results are separated into sections in Table \ref{tbl:result_fastimpute_syn}. The first four sections investigate fastImpute's scalability with respect to each of the 4 parameters $m,n,\mu, k$, with the parameter under investigation denoted in the leftmost column.The final section of the results compares the different algorithms' performance on large realistic combinations of $(m,n,p,k)$.
\begin{table}[t]
\centering
\resizebox{\columnwidth}{!}{%
\begin{tabular}{|c|l|l|l|l||l|c|l|c|l|c|l|c||l|c|}
	\hline \multirow{2}{*}{} & \multirow{2}{*}{$\bm{n}$}&\multirow{2}{*}{$\bm{m}$}&\multirow{2}{*}{$\bm{k}$} & \multirow{2}{*}{$\bm{\mu\%}$} & \multicolumn{2}{c|}{\textbf{fastImpute}} & \multicolumn{2}{c|}{\textbf{SIALS}}& \multicolumn{2}{c|}{\textbf{SISVD}} & \multicolumn{2}{c||}{\textbf{MFSGD}}& \multicolumn{2}{c|}{\textbf{GROUSE}}\\\cline{6-15}
	&  &  & & & $\bm{T}$ & \textbf{MAPE} & $\bm{T}$ & \textbf{MAPE}& $\bm{T}$ & \textbf{MAPE} & $\bm{T}$ & \textbf{MAPE}& $\bm{T}$ & \textbf{MAPE}\\\hline
	\multirow{4}{*}{$n$}&$10^3$ & $10^3$ & 5 & $95\%$ & 0.9s &$3.5\%$ &2.3s & $19.9\%$ &213s & $21.3\%$ &9.4s & $2.1\%$&1.2s & $5.0\%$\\
	&$10^4$ & $10^3$& 5 & $95\%$ & 4.2s&$2.4\%$ & 25.5s&$12.7\%$ &1780s & $17.5\%$ & 72.1s & $2.8\%$& 9.6s & $2.6\%$\\
	&$10^5$ & $10^3$ & 5 & $95\%$ & 18.9s &$2.2\%$ &443s & $8.1\%$&23650s & $12.1\%$ &709s & $6.1\%$&153s & $3.1\%$\\
	&$10^6$ & $10^3$ & 5 & $95\%$ & 104s &$2.1\%$ &6270s & $6.7\%$ &N/A & N/A&7605s & $7.0\%$&1480s & $2.7\%$ \\\hline\hline
	\multirow{4}{*}{$m$}&$10^4$ & $10^3$& 5 & $95\%$ & 4.2s&$2.4\%$ & 25.5s&$12.7\%$ &1780s & $17.5\%$ & 72.1s & $2.8\%$& 9.6s & $2.6\%$\\
	&$10^4$ & $10^4$& 5 & $95\%$ & 19s&$4.0\%$ & 227s &$6.2\%$ &15070s & $8.9\%$ & 840s & $5.4\%$& 75s & $5.1\%$\\
	&$10^4$ & $10^5$ & 5 & $95\%$ & 140s &$3.1\%$ & 3170s & $9.1\%$ &N/A & N/A&8010s & $7.5\%$&748s & $3.6\%$ \\
	&$10^4$ & $10^6$ & 5 & $95\%$ & 1052s &$3.5\%$ & 30542s & $8.0\%$ &N/A & N/A&N/A & N/A&6309s & $3.1\%$\\\hline\hline
	\multirow{4}{*}{$k$}&$10^4$ & $10^3$& 5 & $95\%$ & 4.2s&$2.4\%$ & 25.5s&$12.7\%$ &1780s & $17.5\%$ & 72.1s & $2.8\%$& 9.6s & $2.6\%$\\
	&	$10^4$ & $10^3$ & 10 & $95\%$ & 14.1s&$2.8\%$ &41.0s & $9.8\%$&3120s & $11.7\%$&80.7s & $2.6\%$&17.6s & $2.5\%$\\
&	$10^4$ & $10^3$ & 20 & $95\%$ & 29.8s&$3.7\%$ &80.4s & $8.2\%$&3609s & $14.6\%$&83.7s & $5.9\%$&80.2s & $1.9\%$\\
	&	$10^4$ & $10^3$  & 30 & $95\%$ & 49.5s &$5.0\%$ &122s & $9.3\%$&3670s & $20.9\%$&82.1s & $4.9\%$&196s & $1.1\%$\\\hline\hline
\multirow{4}{*}{$\mu$} &	$10^4$ & $10^3$ & 5 & $20\%$ & 8.4s &$1.1\%$ & 3.6s & $0.6\%$& 204s & $0.6\%$& 140s & $4.9\%$& 21s & $1.9\%$ \\
	&$10^4$ & $10^3$ & 5 & $50\%$ & 7.5s &$1.6\%$ &6.7s & $1.0\%$ &370s & $1.1\%$ &109s & $1.2\%$&15.8s & $2.1\%$\\
	&$10^4$ & $10^3$ & 5 & $80\%$ & 5.7s&$2.4\%$ &13.1s & $4.0\%$&1340s & $2.7\%$ &85s & $1.9\%$&12.5s & $2.4\%$\\
	&$10^4$ & $10^3$& 5 & $95\%$ & 4.2s&$2.6\%$ & 25.5s&$12.7\%$ &1780s & $17.5\%$ & 72.1s & $2.8\%$& 9.6s & $2.6\%$\\\hline\hline
	&$10^4$ & $10^3$& 5 & $95\%$ & 4.2s&$2.5\%$ & 25.5s&$12.7\%$ &1780s & $17.5\%$ & 72.1s & $2.8\%$& 9.6s & $2.6\%$\\
	&$10^5$ & $10^3$& 10 & $95\%$ & 29.0s &$2.5\%$ & 403s & $10.7\%$& 25049s  & $14.6\%$ & 708s & $6.0\%$& 270s & $2.9\%$\\
	&$10^5$ & $10^4$& 10 & $95\%$ & 317s &$2.1\%$ &$4470s$ &$8.9\%$&N/A & N/A &8215s& $7.3\%$&2076s& $2.4\%$\\
	&$10^5$ & $10^5$& 10 & $95\%$ & 3260s &$2.0\%$ &$52690s$ & $4.1\%$ &N/A & N/A  &N/A & N/A&27043s& $2.2\%$ \\
	&$10^6$ & $10^4$& 20 & $95\%$ & 5070s &$1.9\%$ & N/A & N/A & N/A & N/A & N/A & N/A&48740s& $2.9\%$ \\\hline\hline
\end{tabular}
}
\caption{Comparison of fastImpute,  SIALS, SISVD, MFSGD, and GROUSE on synthetic data.  $N/A$ means the algorithm did not complete running in 20 hours, corresponding to 72000 seconds. }
\label{tbl:result_fastimpute_syn}
\end{table}

We see that on the final set of large realistic combinations, fastImpute outperforms all comparison algorithms in all cases. Table \ref{tbl:result_fastimpute_syn_sum} records the average difference in time and MAPE between fastImpute and the other algorithms, on the final set of combinations. On average fastImpute takes $10\%$ of the time of comparison while achieving  $\sim 40-70\%$ reduction in MAPE at the same time. 

\begin{table}[t]
\centering
\resizebox{\columnwidth}{!}{%
\begin{tabular}{|c|l|c|l|c|l|c|l|c|}
\hline
\multirow{2}{*}{}& \multicolumn{2}{c|}{\textbf{SIALS}}& \multicolumn{2}{c|}{\textbf{SISVD}} & \multicolumn{2}{c|}{\textbf{MFSGD}}& \multicolumn{2}{c|}{\textbf{GROUSE}}\\\cline{2-9}
& $\Delta \bm{T}$ & $\Delta$\textbf{MAPE}& $\Delta \bm{T}$ & $\Delta$\textbf{MAPE} &$\Delta \bm{T}$ & $\Delta$\textbf{MAPE}& $\Delta \bm{T}$  & $\Delta$\textbf{MAPE}\\\hline
\textbf{fastImpute vs.}& $-90\%$& $-71\%$ & $-99\%$ & $-95\%$ & $-96\%$ & $-45\%$ & $-81\%$ & $-15\%$ \\\hline
\end{tabular}
}
\caption{Average performance of fastImpute,  SIALS, SISVD, MFSGD and GROUSE on synthetic trials.  Percentages are computed by averaging over the set of realistic combinations.}
	\label{tbl:result_fastimpute_syn_sum}
\end{table}
Even when compared to the online algorithm GROUSE, fastImpute is on average over $80\%$ faster on the final set of combinations.  

For scaling behavior, we have the following observations:
\begin{itemize}
    \item $n,m$ - We see that fastImpute scales sublinearly for low $n,m$ as the gradient descent step dominates, and as we move to $n\sim 10^6$ it starts to scale as $O(nm)$ as the step of completing the final matrix starts to dominate. The MAPE steadily decreases as we have more entries. In contrast, SIALS, SISVD, and MFSGD all roughly scale linearly with $n,m$ from the start. Interestingly, the MFSGD algorithm has increasing error with increasing number of entries - we hypothesize this may be due to the gradient descent in factorized form $\bm{A}=\bm{U}\bm{V}^T$ failing to capture non-linear dynamics of the interactions between $\bm{U}$ and $\bm{V}$ at high levels.
    \item $k$ - Somewhat surprisingly, fastImpute scales as $O(k)$ even though theoretically it scales at $O(k^3)$. We believe this is due to the small constant factor in front of the $k^2$ and $k^3$ terms.
    \item $\mu$ - We see that in accordance to the linear dependence on $|\Omega|$, the number of known entries, fastImpute runs slower as we have more filled elements, while in contrast SIALS and SISVD both run faster. MFSGD is similar to fastImpute in that it runs slower with more filled elements (a construct of the gradient descent method). 
\end{itemize}
\subsection{Real-World Experiments}
For real-world experiments, we  again utilize the Netflix datasets created in Section \ref{sec:fastImputeSrealexp} without the feature matrix. 

We test against SIALS as it is the only algorithm capable of scaling to such size. We split the training set in $80\%/20\%$, where the latter group is used for validation of the rank in SIALS and fastImpute. We then report the time taken, $T$, the MAPE, and the optimal chosen rank $k^*$ for each algorithm:
\begin{table}[t]
	\centering
	\resizebox{\columnwidth}{!}{%
	\begin{tabular}{|l|l|l||l|l|l|l|l|l|}
		\hline  \multirow{2}{*}{$\bm{n}$}&\multirow{2}{*}{$\bm{m}$} & \multirow{2}{*}{$\bm{\mu\%}$} & \multicolumn{3}{c|}{\textbf{fastImpute}} & \multicolumn{3}{c|}{\textbf{SIALS}}\\\cline{4-9}
		 &  &    & $\bm{T}$  & $\bm{k^*}$ & \textbf{MAPE}  & $\bm{T}$  & $\bm{k^*}$ & \textbf{MAPE} \\\hline
			3,923 & 103  & $92.6\%$ & 2s & 5 & $23.5\%$ &  4s & 5 & $30.6\%$ \\
		18,227 & 323   & $94.8\%$ & 18s & 8 & $19.8\%$ & 47s & 7 & $27.5\%$\\
		96,601 & 788  & $94.2\%$ & 109s & 10 & $18.2\%$ & 620s & 10 & $24.0\%$\\
		471,268 & 1,760  & $93.6\%$ & 370s & 10 & $16.5\%$ & 2837s & 12 & $22.5\%$ \\
		471,268 & 14,538 & $94.1\%$ & 2098s & 12 & $13.8\%$ & 38256s & 14 & $20.1\%$\\\hline
	\end{tabular}
	}
	\caption{Comparison of methods on Netflix data for fastImpute}
\end{table}
We see that fastImpute is able to outperform SIALS on the Netflix dataset across the different $n$ and $m$ values, while enjoying superior scalability especially as we approach the full matrix. 
\section{Conclusion}
\label{sec:mc_conclusion}
In conclusion, we have designed a unified optimization framework that is able to conduct state-of-the-art matrix completion with and without side information. Using the factorization approach $\bm{A}=\bm{U}\bm{S}^T\bm{B}^T$, we wrote $\bm{U}=f(\bm{S})$ as a function of $\bm{S}$, and derived the cost and gradient expressions with respect to $\bm{S}$ through a separable reformulation of the problem. By then conducting non-convex gradient descent on $\bm{S}$, our synthetic and real-world data experiments show the competitiveness of the method in both scalability and accuracy against a multitude of comparison algorithms.
\vfill
\pagebreak
\bibliography{fastImpute}
\clearpage
\appendix

\setcounter{table}{0}
\renewcommand{\thetable}{A\arabic{table}}
\setcounter{figure}{0}
\renewcommand{\thefigure}{A\arabic{figure}}
\setcounter{equation}{0}
\renewcommand{\theequation}{A\arabic{equation}}
\section{Proof of Theorem \ref{theo:localmin}}
\label{app:localminproof}
For the case where we have no side information, our objective function is:
\[c(\bm{S})= \frac{1}{nm}\sum_{i=1}^n  \overline{\bm{a}}_i
\left(\bm{I}_m-\bm{S}\left(\bm{S}^T\bm{W}_i\bm{S}\right)^{-1}\bm{S}^T\right)\overline{\bm{a}}_i^T.\]

We  first outline the roadmap to prove the two results required. As a reminder, a function $f: \md{R}^d \to \md{R}$  is $(\theta, \zeta, \eta)$-strict saddle if for every $\bm{x}$, at least one of the following hold for constants $\theta, \zeta, \eta>0$:
\begin{enumerate}
	\item $\|\nabla f(\bm{x})\|\geq \theta$.
	\item $\lambda_{\text{min}}(\nabla^2f(\bm{x}))\leq -\zeta$.
	\item $\bm{x}$ is $\eta$-close to a local minimum.
\end{enumerate}
We want to prove that our objective function $c(\bm{S})$ is a $(\theta, \zeta, \eta)$-strict saddle. Furthermore, we want to prove that there are no spurious local minima. 

To do so, we  claim that proving the following statement would imply both results.

``For all $\bm{S}$ such that $\|\nabla c(\bm{S})\|\leq \epsilon$ and $\bm{S}$ more than $\eta$-away from a \emph{global} minimum, we have that $\lambda_{\text{min}}(\nabla^2f(\bm{x}))\leq -\zeta$ for some suitable $\zeta$."

By construction, this statement shows that for $c(\bm{S})$, if Condition 1 and 3 of the strict saddle definition is not true, then 2 is always true, which implies $c(\bm{S})$ is $(\theta, \zeta, \eta)$-strict saddle. 

Furthermore, this statement implies that every stationary point (which satisfies $\|\nabla c(\bm{S})\|\leq \epsilon$) that is not a global minimum has a negative eigenvalue in its Hessian. Thus, every stationary point that is not a global minimum is a saddle point, since local minima have no negative eigenvalues in the Hessian. Therefore, there are no spurious local minima - every local minimum is a global minimum. 

Thus, proving this statement would give us the two required results we need. Therefore, we now set out to prove the statement above. 

To ease the notation burden, we  define the following :
\begin{align*}
\bm{S}_i&=\bm{W}_i\bm{S},\\
\bm{P}_{\bm{S}}&=\bm{I}_m - \bm{S}(\bm{S}^T\bm{S})^{-1}\bm{S}^T,\\
\bm{P}_{\bm{S}_i}&=\bm{W}_i - \bm{S}_i(\bm{S}_i^T\bm{S}_i)^{-1}\bm{S}_i^T.
\end{align*}
Note that the projection matrices are setup so that $\bm{P}_{\bm{S}_i}\bm{S}_i=\bm{0}$ and $\bm{P}_{\bm{S}}\bm{S}=\bm{0}$. Furthermore, let us denote $\bm{U}^*\in \md{R}^{n \times k},\bm{S}^*\in \md{R}^{m \times k}$ as the true solution to the matrix completion problem (i.e., $\bm{A}=\bm{U}^*\bm{S}^{*T}$). Then we can write 
\begin{equation}
c(\bm{S})= \frac{1}{nm}\sum_{i=1}^n  \bm{a}_i\bm{P}_{\bm{S}_i}\overline{\bm{a}}_i^T. \label{eq:objective} 
\end{equation}
We calculate its gradient  as follows.
\begin{lem}
	\label{lem:first_deriv}
	\begin{align*}
	\nabla c(\bm{S}) = \frac{1}{nm}\sum_{i=1}^n  \bm{P}_{\bm{S}_i}\bm{a}_i^T\bm{a}_i\bm{S}_i(\bm{S}_i^T\bm{S}_i)^{-1}.
	\end{align*}
\end{lem}
\begin{proof}
	This follows from direct calculation of the derivative. In particular, we note that $\bm{S}^T\bm{P}_{\bm{S}_i}=\bm{0}$ for all $i$ by definition, and thus \emph{every term} of the derivative in the sum is orthogonal to $\bm{S}$. 
\end{proof}
To calculate the Hessian, note that we expressed   $\bm{S}$ as a $m\times k$ matrix, so the Hessian has  dimensions $(m\times k) \times (m \times k)$. Therefore, to calculate the second derivative in a specific direction $\bm{M}$, we   use  the following notation  to represent the quadratic form: 
\[\bm{M} : \nabla ^2 c(\bm{S}): \bm{M}=\sum_{i,j,k,l} M_{ij}\nabla^2 c(\bm{S})_{ijkl}M_{kl}.\] 
Where $\nabla^2 c(\bm{S})_{ijkl}=\frac{\partial^2 c(\bm{S})}{\partial \bm{S}_{ij} \partial \bm{S}_{kl}}$.    
\begin{lem}
	\label{lem:second_deriv}
	For any matrix $\bm{M}\in \md{R}^{m \times k}$, the Hessian in the direction of $\bm{M}$ is
	\begin{align*}
	\bm{M} : \nabla ^2 c(\bm{S}): \bm{M}= \frac{2}{nm} \sum_{i=1}^n &\biggl( \bm{a}_i\bm{S}_i(\bm{S}_i^T\bm{S}_i)^{-1}\bm{M}^T\bm{P}_{\bm{S}_i}\bm{M} (\bm{S}_i^T\bm{S}_i)^{-1}\bm{a}_i^T\biggr.\\&+2\bm{a}_i\bm{S}_i(\bm{S}_i^T\bm{S}_i)^{-1}\bm{M}^T\bm{S}_i(\bm{S}_i^T\bm{S}_i)^{-1}\bm{M}^T\bm{P}_{\bm{S}_i}\bm{a}_i^T\\& -  \overline{\bm{a}}_i \bm{M}(\bm{S}_i^T\bm{S}_i)^{-1}\bm{M}^T \bm{P}_{\bm{S}_i} \bm{a}_i^T\\&+\biggl.\bm{a}_i\bm{S}_i(\bm{S}_i^T\bm{S}_i)^{-1} \bm{S}^T_i\bm{M}(\bm{S}_i^T\bm{S}_i)^{-1}\bm{M}^T\bm{P}_{\bm{S}_i}\bm{a}_i^T\biggr).
	\end{align*}
\end{lem}
\begin{proof}
	The result follows through by direct calculation. 
\end{proof}


As a reminder, our goal is to prove that, given $\|\nabla c(\bm{S})\|\leq \epsilon$ and $\|\bm{S}-\bm{S}^*\|\geq \eta$ for any global minima $\bm{S}^*$, there exists a negative eigenvalue for the Hessian $\nabla^2 c(\bm{S})$.

Note the existence of a negative eigenvalue for the Hessian $\nabla^2 c(\bm{S})$ is equivalent to existence of a matrix $\bm{M}$ such that $\bm{M} : \nabla ^2 c(\bm{S}): \bm{M}<0$.  Now let us look at the expression for the second derivative.
\begin{align*}
\bm{M} : \nabla ^2 c(\bm{S}): \bm{M}= \frac{2}{nm} \sum_{i=1}^n &\biggl( \underbrace{\bm{a}_i\bm{S}_i(\bm{S}_i^T\bm{S}_i)^{-1}\bm{M}^T\bm{P}_{\bm{S}_i}\bm{M} (\bm{S}_i^T\bm{S}_i)^{-1}\bm{S}_i^T\bm{a}_i^T}_{\mytag{(T1)}{term1}}\biggr.\\&+\underbrace{2\bm{a}_i\bm{S}_i(\bm{S}_i^T\bm{S}_i)^{-1}\bm{M}^T\bm{S}_i(\bm{S}_i^T\bm{S}_i)^{-1}\bm{M}^T\bm{P}_{\bm{S}_i}\bm{a}_i^T}_{\mytag{(T2)}{term2}}\\& -  \underbrace{\overline{\bm{a}}_i \bm{M}(\bm{S}_i^T\bm{S}_i)^{-1} \bm{M}^T\bm{P}_{\bm{S}_i} \bm{a}_i^T}_{\mytag{(T3)}{term3}}\\&+\biggl.\underbrace{\bm{a}_i\bm{S}_i(\bm{S}_i^T\bm{S}_i)^{-1} \bm{S}^T_i\bm{M}(\bm{S}_i^T\bm{S}_i)^{-1}\bm{M}^T \bm{P}_{\bm{S}_i}\bm{a}_i^T}_{\mytag{(T4)}{term4}}\biggr).
\end{align*}
We first prove that for \emph{any} $\bm{M}$, the terms \ref{term2} and  \ref{term4} are small. Then,   we   exhibit a specific $\bm{M}$, such that \ref{term1} is close to 0, \ref{term3} is large and positive, and so that the resulting quadratic form $\bm{M} : \nabla ^2 c(\bm{S}): \bm{M}$ is negative. 

Now let us prove that terms  \ref{term2} and  \ref{term4} are small (on the order of $O(\epsilon)$) for any $\bm{M}$.

\begin{lem}
	\label{lem:term_24_bound}
	Let $\bm{S}$ satisfy Assumptions \ref{ass:ripweak}-\ref{ass:Vrestrict}, and be such that $\|\nabla c(\bm{S})\|\leq \epsilon$. Then we have that, for $r>\frac{Ck}{m}$ for sufficiently large $C$, with probability at least $1-O(\frac{1}{n})$:
	\begin{align*}
	\left\|\frac{1}{nm}\sum_{i=1}^n\bm{a}_i\bm{S}_i(\bm{S}_i^T\bm{S}_i)^{-1}\bm{M}^T\bm{S}_i(\bm{S}_i^T\bm{S}_i)^{-1}\bm{M}^T\bm{P}_{\bm{S}_i}\bm{a}_i^T\right\|&\leq Kr\epsilon\|\bm{M}\|^2,\\
	\left\|\frac{1}{nm}\sum_{i=1}^n\bm{a}_i\bm{S}_i(\bm{S}_i^T\bm{S}_i)^{-1} \bm{S}^T_i\bm{M}(\bm{S}_i^T\bm{S}_i)^{-1}\bm{M}^T \bm{P}_{\bm{S}_i}\bm{a}_i^T\right\|&\leq Lr\epsilon\|\bm{M}\|^2 ,
	\end{align*}
	for some absolute constants $K,L$. 
\end{lem}
\begin{proof}
	Note that we are trying to bound the second derivative when the first derivative is approximately 0. Using Lemma \ref{lem:first_deriv}, we can rewrite $\|\nabla c(\bm{S})\|\leq \epsilon$ as
	\begin{equation}
	\left\|\frac{1}{nm}\sum_{i=1}^n \bm{a}_i\bm{S}_i(\bm{S}_i^T\bm{S_i})^{-1}\bm{M}^T\bm{P}_{\bm{S}_i} \bm{a}_i^T\right\|\leq \epsilon\|\bm{M}\|. \label{eq:first_deriv_alt}
	\end{equation}
	We  only prove for term \ref{term2} as the  proof for  term \ref{term4} follows the same steps. First, we 
	bound the difference between the first derivative in \eqref{eq:first_deriv_alt} and its ``non-stochastic" version:
	\begin{equation}
	\left\|\frac{1}{nm} \sum_{i=1}^n \bm{a}_i\bm{S}_i(\bm{S}_i^T\bm{S}_i)^{-1}\bm{M}^T\bm{P}_{\bm{S}_i}\bm{a}_i^T -\frac{r}{nm} \sum_{i=1}^n \bm{a}_i\bm{S}(\bm{S}^T\bm{S})^{-1}\bm{M}^T\bm{P}_{\bm{S}}\bm{a}_i^T\right\|. \label{eq:first_deriv_diff}
	\end{equation}
	Then we focus on each individual term inside the sum (taking $\frac{1}{m}$ inside). We have
	\begin{align*}
	&\left\|\frac{\bm{a}_i\bm{S}_i(\bm{S}_i^T\bm{S}_i)^{-1}\bm{M}^T\bm{P}_{\bm{S}_i}\bm{a}_i^T}{m}-\frac{l}{m}\bm{a}_i\bm{S}(\bm{S}^T\bm{S})^{-1}\bm{M}^T\bm{P}_{\bm{S}}\bm{a}_i^T\right\|\\&\leq r\left\|\left(\frac{\bm{a}_i\bm{S}_i}{l}-\frac{\bm{a}_i\bm{S}}{m}\right)\left(\frac{\bm{S}_i^T\bm{S}_i}{l}\right)^{-1}\frac{\bm{M}^T\bm{P}_{\bm{S}_i}\bm{a}_i^T}{l}\right\|\\&+r\left\|\frac{\bm{a}_i\bm{S}}{m}\left(\left(\frac{\bm{S}_i^T\bm{S}_i}{l}\right)^{-1}-\left(\frac{\bm{S}^T\bm{S}}{m}\right)^{-1}\right)\frac{\bm{M}^T\bm{P}_{\bm{S}_i}\bm{a}_i^T}{l}\right\|\\&+r\left\|\frac{\bm{a}_i\bm{S}}{m}\left(\frac{\bm{S}^T\bm{S}}{m}\right)^{-1}\left(\frac{\bm{M}^T\bm{P}_{\bm{S}_i}\bm{a}_i^T}{l}-\frac{\bm{M}^T\bm{P}_{\bm{S}}\bm{a}_i^T}{m}\right)\right\|.
	\end{align*}
	
	We  utilize Hoeffding's theorem for the first and third term, along with Lemma \ref{lem:matrix_inv_ineq} for the second term to obtain 
	that with probability $1-\delta$
	the sum of the three terms is less than or equal to $r\sqrt{\frac{D k\log(\frac{1}{\delta})}{l}}\|\bm{M}\|,$
	where $D$ is an absolute  constant.

	Then, we have, by Lemma \ref{lem:hoeffding_general}, with probability $1-\delta$
	\begin{align*}
	&\left\|\frac{1}{nm} \sum_{i=1}^n \bm{a}_i\bm{S}_i(\bm{S}_i^T\bm{S}_i)^{-1}\bm{M}^T\bm{P}_{\bm{S}_i}\bm{a}_i^T -\frac{l}{nm} \sum_{i=1}^n \bm{a}_i\bm{S}(\bm{S}^T\bm{S})^{-1}\bm{M}^T\bm{P}_{\bm{S}}\bm{a}_i^T\right\|\\&\leq r\sqrt{\frac{D' k\log(\frac{1}{\delta})}{ln}}\|\bm{M}\|,
	\end{align*}
	where $D'$ is an absolute  constant. 
	We let  $\delta =\frac{1}{n}$, and $l\geq Ck$ for some constant $C$. Then, for $n$ sufficiently large, we have
	$$r\sqrt{\frac{D' k\log(\frac{1}{\delta})}{ln}}\|\bm{M}\| \leq r\epsilon \|\bm{M}\|.$$
	
	Therefore, we know that, with probability at least $1-\frac{1}{n}$, we have 
	\begin{equation}
	\left\|\frac{l}{nm} \sum_{i=1}^n \bm{a}_i\bm{S}(\bm{S}^T\bm{S})^{-1}\bm{M}^T\bm{P}_{\bm{S}}\bm{a}_i^T\right\|\leq 2r\epsilon \|\bm{M}\|, \label{eq:first_deriv_bound}
	\end{equation}
	for any $\bm{M}$. Now we 
	do the same for term \ref{term2} by bounding the difference between term \ref{term2} and its non-stochastic variant. The details are omitted as they follow the same logic as above. The final result is that with probability at least $1-\frac{1}{n}$, we have 
	\begin{align}
	&\|\frac{1}{nm}\sum_{i=1}^n\bm{a}_i\bm{S}_i(\bm{S}_i^T\bm{S}_i)^{-1}\bm{M}^T\bm{S}_i(\bm{S}_i^T\bm{S}_i)^{-1}\bm{M}^T\bm{P}_{\bm{S}_i}\bm{a}_i^T\nonumber \\&-\underbrace{\frac{l}{nm}\sum_{i=1}^n\bm{a}_i\bm{S}(\bm{S}^T\bm{S})^{-1}\bm{M}^T\bm{S}(\bm{S}^T\bm{S})^{-1}\bm{M}^T\bm{P}_{\bm{S}}\bm{a}_i^T}_{\mytag{(S1)}{expr1}}\|\leq r\epsilon \|\bm{M}\|^2.\label{eq:term_2_diff}
	\end{align}
	Now let us focus on the term \ref{expr1}. 
	Such term is exactly (\ref{eq:first_deriv_bound}) substituting $\bm{M}$ with  $\bm{M}^T\bm{S}(\bm{S}^T\bm{S})^{-1}\bm{M}^T$. Therefore, we have 
	\begin{equation}
	\left\|\frac{l}{nm}\sum_{i=1}^n\bm{a}_i\bm{S}(\bm{S}^T\bm{S})^{-1}\bm{M}^T\bm{S}(\bm{S}^T\bm{S})^{-1}\bm{M}^T\bm{P}_{\bm{S}}\bm{a}_i^T \right\|\leq Kr\epsilon\|\bm{M}\|^2, \label{eq:term_2_alt_bound}
	\end{equation}
	for some constant $K$. Then combining (\ref{eq:term_2_diff}) and (\ref{eq:term_2_alt_bound}) gives the required result. 
\end{proof}

We next  utilize the bounds on   \ref{term2} and \ref{term4}
to explicitly construct a $\bm{M}$ to bound \ref{term1} and \ref{term3}. Specifically, we define $\bm{M}=\bm{z}^T\bm{x}$, $\|\bm{M}\|=1$,  where $\bm{z}^T \in \md{R}^{m \times 1}$ is a non-zero (rescaled) column of $\bm{P}_{\bm{S}}\bm{S}^*$ (it exists because $\bm{S}$ is not optimal, so $\bm{P}_{\bm{S}}\bm{S}^*\neq \bm{0}$) and $\bm{x}^T \in \md{R}^{k \times 1}$ is an eigenvector of $\bm{S}^{*T}\bm{S}(\bm{S}^T\bm{S})^{-1}$ that has an eigenvalue of size $O(\epsilon)$. We utilize the row vector convention to be consistent with our other notations. We prove the existence of the eigenvector $\bm{x}^T$ in Lemma \ref{lem:eigvec_existence} below.
\begin{lem}
	\label{lem:eigvec_existence}
	Let $\bm{S}$ satisfy Assumptions \ref{ass:ripweak}-\ref{ass:Vrestrict}, and be such that $\|\nabla c(\bm{S})\|\leq \epsilon$ for some $\epsilon$ sufficiently small. Assume that $\min_{\bm{R}^T\bm{R}=\bm{I}}\|\bm{S}-\bm{S}^*\bm{R}\|\geq \eta$. Then $\bm{S}^{*T}\bm{S}(\bm{S}^T\bm{S})^{-1}$ has an eigenvector with an eigenvalue of size $O(\epsilon)$.
\end{lem}
\begin{proof}
	First it is easy to see that
	\begin{equation}
	\min_{\bm{R}^T\bm{R}=\bm{I}}\|\bm{S}^*-\bm{S}\bm{R}\|  =\|\bm{P}_{\bm{S}}\bm{S}^*\|\geq \eta. \label{eq:nonoptimality}
	\end{equation}
	Now, let us consider the first derivative bound from (\ref{eq:first_deriv_bound}):
	\begin{equation*}
	\left\|\frac{1}{nm} \sum_{i=1}^n \bm{a}_i\bm{S}(\bm{S}^T\bm{S})^{-1}\bm{M}^T\bm{P}_{\bm{S}}\bm{a}_i^T\right\|\leq 2r\epsilon \|\bm{M}\|.
	\end{equation*}
	We  prove that given this, there exists an eigenvector of $\bm{S}^{*T}\bm{S}(\bm{S}^T\bm{S})^{-1}$ that has an eigenvalue of size $O(\epsilon)$. Assume that there is not. 
	By definition $\bm{a}_i=\bm{u}_i^*\bm{S}^{*T}$, so we can write the bound as:
	\begin{equation}
	\left\|\frac{1}{nm} \sum_{i=1}^n \bm{u}_i^*\bm{S}^{*T}\bm{S}(\bm{S}^T\bm{S})^{-1}\bm{M}^T\bm{P}_{\bm{S}}\bm{S}^{*}\bm{u}_i^{*T}\right\|\leq 2r\epsilon \|\bm{M}\|.\label{eq:first_deriv_term_1}
	\end{equation}
	Then since there is not an eigenvalue of magnitude $O(\epsilon)$, let the smallest eigenvalue of $\bm{S}^{*T}\bm{S}(\bm{S}^T\bm{S})^{-1}$ have a magnitude of $O(\gamma)\gg O(\epsilon)$. Let $\bm{x}^T \in \md{R}^{k \times 1}$ be the eigenvector associated with such eigenvalue. Then $\bm{S}^{*T}\bm{S}(\bm{S}^T\bm{S})^{-1}\bm{x}^T=O(\gamma)\bm{x}^T$.   From Equation (\ref{eq:nonoptimality}) we know that $\|\bm{P}_{\bm{S}}\bm{S}^*\|\geq \eta$, so let $\bm{y}\in \md{R}^{1 \times m}$ be such that $\bm{y}\bm{P}_{\bm{S}}\bm{S}^*=O(\eta)\bm{v}^T$ where $\bm{v} \in \md{R}^{1 \times k}$, and rescaled such that for $\bm{M}=\bm{y}^T\bm{x}$, we have $\|\bm{M}\|=O(1)$. Then we have
	\begin{equation*}
	\left\|\frac{1}{nm} \sum_{i=1}^n \bm{u}_i^*\bm{S}^{*T}\bm{S}(\bm{S}^T\bm{S})^{-1}\bm{M}^T\bm{P}_{\bm{S}}\bm{S}^{*}\bm{u}_i^{*T}\right\|= \left\|\frac{O(\eta\gamma)}{nm} \sum_{i=1}^n \bm{u}_i^*\bm{x}\bm{v}^T\bm{u}_i^{*T}\right\|=O\left(\frac{\eta\gamma}{m}\right)\leq O(\epsilon),
	\end{equation*}
	where the last equality follows from the submatrix full-rank property of $\bm{U}^*$ (note that $\bm{u}_i^* \in \md{R}^{1 \times k}$ are row vectors here). Since $\eta$ is independent from $\epsilon$ ($\eta$ only depends on how far $\bm{S}$ is from $\bm{S}^*$) and $\gamma \gg \epsilon$ by setup, we obtain    a contradiction for sufficiently small $\epsilon$.
	
	Therefore, there exists $\epsilon$ sufficiently small such that, there is an eigenvector of $\bm{S}^{*T}\bm{S}(\bm{S}^T\bm{S})^{-1}$, $\bm{x}^T$, that has an eigenvalue of size $O(\epsilon)$.
\end{proof}
Having proved that the desired matrix $\bm{M}$ does indeed exist, we   show that indeed with such $\bm{M}$, \ref{term1} is small, and \ref{term3} is large and positive.
We first bound \ref{term1}.
\begin{lem}
	\label{lem:term_1_bound}
	Let $\bm{S}$ satisfy Assumptions \ref{ass:ripweak}-\ref{ass:Vrestrict}, and be such that $\|\nabla c(\bm{S})\|\leq \epsilon$ for some $\epsilon$ sufficiently small. Assume that $\min_{\bm{R}^T\bm{R}=\bm{I}}\|\bm{S}-\bm{S}^*\bm{R}\|\geq \eta$. Then for $\bm{M}=\bm{z}^T\bm{x}$ where $\bm{z}$, $\bm{x}$ defined above, with probability at least $1-O(\frac{1}{n})$:
	\begin{align*}
	\left\|\frac{1}{nm}\sum_{i=1}^n \bm{a}_i\bm{S}_i(\bm{S}_i^T\bm{S}_i)^{-1}\bm{M}^T\bm{P}_{\bm{S}_i}\bm{M} (\bm{S}_i^T\bm{S}_i)^{-1}\bm{S}_i^T\bm{a}_i^T \right\|\leq O(r\epsilon),
	\end{align*}
\end{lem}
\begin{proof}
	From Lemma \ref{lem:eigvec_existence}, we know that there is an eigenvector of $\bm{S}^{*T}\bm{S}(\bm{S}^T\bm{S})^{-1}$, $\bm{x}^T$, that has an eigenvalue of size $O(\epsilon)$.
	Similar to the proof for Lemma \ref{lem:term_24_bound}, we can bound term \ref{term1} with the non-stochastic version with at least probability $1-\frac{1}{n}$:
	\begin{align}
	&\left\|\frac{1}{nm}\sum_{i=1}^n \bm{a}_i\bm{S}_i(\bm{S}_i^T\bm{S}_i)^{-1}\bm{M}^T\bm{P}_{\bm{S}_i}\bm{M} (\bm{S}_i^T\bm{S}_i)^{-1}\bm{S}_i^T\bm{a}_i^T \right.\nonumber \\&\left. -\underbrace{\frac{r}{n}\sum_{i=1}^n \bm{a}_i\bm{S}(\bm{S}^T\bm{S})^{-1}\bm{M}^T\bm{P}_{\bm{S}}\bm{M} (\bm{S}^T\bm{S})^{-1}\bm{S}^T\bm{a}_i^T}_{\mytag{(S2)}{expr2}} \right\|\leq r\epsilon \|\bm{M}\|^2.  \label{eq:term_1_diff}
	\end{align}
	Let us look at each term in the sum \ref{expr2}. 
	Taking $\bm{M}=\bm{z}^T\bm{x}$. We have
	\begin{align*}
	&\|\bm{a}_i\bm{S}(\bm{S}^T\bm{S})^{-1}\bm{M}^T\bm{P}_{\bm{S}}\bm{M} (\bm{S}^T\bm{S})^{-1}\bm{S}^T\bm{a}_i^T\|\\&= \|\bm{P}_{\bm{S}}\bm{M} (\bm{S}^T\bm{S})^{-1}\bm{S}^T\bm{a}_i^T\|^2\\& = \|\bm{P}_{\bm{S}}\bm{M} (\bm{S}^T\bm{S})^{-1}\bm{S}^T\bm{S}^{*}\bm{u}_i^{*T}\|^2\\& =\|\bm{P}_{\bm{S}}\bm{z}^T\bm{x}(\bm{S}^T\bm{S})^{-1}\bm{S}^T\bm{S}^{*}\bm{u}_i^{*T}\|^2 \\& = O(\epsilon^2) \|\bm{z}^T\bm{x}\bm{u}_i^{*T}\|^2\\& = O(\epsilon^2),
	\end{align*}
	where in the last equality we used the identity that $\bm{P}_{\bm{S}}\bm{z}^T=\bm{z}^T$. Then, we can bound the sum \ref{expr2} as
	\begin{align*}
	&\left\|\frac{r}{n}\sum_{i=1}^n \bm{a}_i\bm{S}(\bm{S}^T\bm{S})^{-1}\bm{M}^T\bm{P}_{\bm{S}}\bm{M} (\bm{S}^T\bm{S})^{-1}\bm{S}^T\bm{a}_i^T\right\|\\&\leq \frac{r}{n}\sum_{i=1}^n \|\bm{a}_i\bm{S}(\bm{S}^T\bm{S})^{-1}\bm{M}^T\bm{P}_{\bm{S}}\bm{M} (\bm{S}^T\bm{S})^{-1}\bm{S}^T\bm{a}_i^T\|\\&=O(r\epsilon^2).
	\end{align*}
	Substituting such bound on equation \ref{expr2} into (\ref{eq:term_1_diff}), we have that
	\[\left\|\frac{1}{nm}\sum_{i=1}^n \bm{a}_i\bm{S}_i(\bm{S}_i^T\bm{S}_i)^{-1}\bm{M}^T\bm{P}_{\bm{S}_i}\bm{M} (\bm{S}_i^T\bm{S}_i)^{-1}\bm{S}_i^T\bm{a}_i^T\right\|\leq O(r\epsilon).\]
	As here we have $\|\bm{M}\|=O(1)$. 
\end{proof}
Now, finally, we prove that \ref{term3} is large under such $\bm{M}$.
\begin{lem}
	\label{lem:term_3_bound}
	Let $\bm{S}$ satisfy Assumptions \ref{ass:ripweak}-\ref{ass:Vrestrict}. Further assume that $\displaystyle \min_{\bm{R}^T\bm{R}=\bm{I}} \|\bm{S}^*-\bm{S}\bm{R}\|\geq \eta$ and $\|\nabla c(\bm{S})\|\leq \epsilon$  for some $\epsilon$ sufficiently small. Then for $\bm{M}=\bm{z}^T\bm{x}$ where $\bm{z}$, $\bm{x}$ defined above, with probability at least $1-O(\frac{1}{n})$:
	\begin{align*}
	\frac{1}{nm}\sum_{i=1}^n \bm{a}_i \bm{M}(\bm{S}_i^T\bm{S}_i)^{-1} \bm{M}^T\bm{P}_{\bm{S}_i} \bm{a}_i^T \geq O\left(\frac{r\eta^2}{k}\right),
	\end{align*}
\end{lem}
\begin{proof}
	First it is easy to see that
	\begin{equation}
	\min_{\bm{R}^T\bm{R}=\bm{I}}\|\bm{S}^*-\bm{S}\bm{R}\|  =\|\bm{P}_{\bm{S}}\bm{S}^*\|\geq \eta.
	\end{equation}
	We 
	consider the term $\frac{1}{nm}\sum_{i=1}^n \bm{a}_i \bm{M}(\bm{S}^T\bm{S})^{-1} \bm{M}^T\bm{P}_{\bm{S}} \bm{a}_i^T$. We have that
	\begin{align*}
	&\frac{l}{nm}\sum_{i=1}^n \bm{a}_i \bm{M}(\bm{S}^T\bm{S})^{-1} \bm{M}^T\bm{P}_{\bm{S}} \bm{a}_i^T\\&= \frac{l}{nm}\sum_{i=1}^n \bm{a}_i \bm{z}^T\bm{x} (\bm{S}^T\bm{S})^{-1} \bm{x}^T\bm{z}\bm{a}_i^T ,
	\end{align*}
	where we utilized $\bm{z}\bm{P}_{\bm{S}}=\bm{z}$. By definition $\bm{a}_i^T=\bm{S}^{*}\bm{u}_i^{*T}$, so we have
	$$ \frac{l}{nm}\sum_{i=1}^n \bm{a}_i \bm{z}^T\bm{x} (\bm{S}^T\bm{S})^{-1} \bm{x}^T\bm{z}\bm{a}_i^T 
	=\frac{r}{n}\sum_{i=1}^n \bm{u}_i^{*}\bm{S}^{*T} \bm{z}^T\bm{x}(\bm{S}^T\bm{S})^{-1} \bm{x}^T\bm{z}\bm{S}^{*}\bm{u}_i^{*T}.$$
	
	Since $(\bm{S}^T\bm{S})^{-1}$ is positive definite (PD), we have a sum of quadratic forms over a PD matrix, which is positive as long as $\bm{x}^T\bm{z}\bm{S}^{*}\bm{u}_i^{*T}$ is non-zero. Since we have that $\bm{z}$ is a non-zero column vector of $\bm{P}_{\bm{S}}\bm{S}^{*}$, and we know that $\|(\bm{P}_{\bm{S}}\bm{S}^{*})^T\bm{S}^{*}\|=\|\bm{S}^{*T}\bm{P}_{\bm{S}}\bm{S}^{*}\|^2\geq O(\eta^2)$, so in particular, we have that $\|\bm{x}^T\bm{z}\bm{S}^{*}\|\geq O(\eta^2)$. Thus, we only need $\bm{u}_i^{*T}$ to not always be in the null space of $\bm{x}^T\bm{z}\bm{S}^{*}$. Now by the sub-matrix full rank condition of $\bm{A}$, we know that $\bm{u}_i^*$ is not in the null space of $\bm{x}^T\bm{z}\bm{S}^{*}$ for $O(n)$ indices. Therefore, we have that $\|\bm{x}^T\bm{z}\bm{S}^{*}\bm{u}_i^{*T}\|=O(\eta^2)$. Then, we have 
	$$\frac{r}{n}\sum_{i=1}^n \bm{u}_i^{*}\bm{S}^{*T} \bm{z}^T\bm{x}(\bm{S}^T\bm{S})^{-1} \bm{x}^T\bm{z}\bm{S}^{*}\bm{u}_i^{*T}
	\geq \frac{r}{\sigma_{max}(\bm{S})} C\eta^2,$$
	for some constant $C$ and where $\sigma_{max}(\bm{S})$ is the largest singular value of $\bm{S}$. By the nuclear-Frobenius norm inequality, we have \[\|\bm{S}^T\bm{S}\|_*\leq \rank(\bm{S}^T\bm{S})\|\bm{S}\|^2=k.\]
	Therefore, the maximum singular value of $\bm{S}$ is at most $k$, resulting in the bound
	\begin{equation}   \frac{r}{\sigma_{max}(\bm{S})} C\eta^2 \geq  \frac{Cr\eta^2}{k}.
	\label{eq:temp34}
	\end{equation}
	
	By a similar proof to what is done in Lemma \ref{lem:term_24_bound}, we can show that with probability greater than $1- O(\frac{1}{n})$ we have
	\begin{equation}
	\left\|\frac{1}{nm}\sum_{i=1}^n \bm{a}_i \bm{P}_{\bm{S}}\bm{S}^*(\bm{S}^T\bm{S})^{-1} \bm{S}^{*T}\bm{P}_{\bm{S}}\bm{a}_i^T -\frac{l}{nm}\sum_{i=1}^n \overline{\bm{a}}_i \bm{P}_{\bm{S}}\bm{S}^*(\bm{S}^T\bm{S})^{-1} \bm{S}^{*T}\bm{P}_{\bm{S}}\bm{P}_{\bm{S}_i}\bm{a}_i^T \right\|\leq \frac{Cr\eta^2}{2k}. \label{eq:term_3_diff}
	\end{equation}
	Combining (\ref{eq:term_3_diff}) with \eqref{eq:temp34} proves  the lemma.
\end{proof}
Finally, we combine the results of  Lemmas \ref{lem:term_24_bound}, \ref{lem:term_1_bound} and \ref{lem:term_3_bound} to prove that for $\bm{M}=\bm{z}^T\bm{x}$, $\|\bm{M}\|=1$ defined previously, we have
\begin{align*}
\bm{M} : \nabla ^2 c(\bm{S}): \bm{M}&= \frac{2}{nm} \sum_{i=1}^n \biggl( \bm{a}_i\bm{S}_i(\bm{S}_i^T\bm{S}_i)^{-1}\bm{M}^T\bm{P}_{\bm{S}_i}\bm{M} (\bm{S}_i^T\bm{S}_i)^{-1}\bm{a}_i^T\biggr.\\&+2\bm{a}_i\bm{S}_i(\bm{S}_i^T\bm{S}_i)^{-1}\bm{M}^T\bm{S}_i(\bm{S}_i^T\bm{S}_i)^{-1}\bm{M}^T\bm{P}_{\bm{S}_i}\bm{a}_i^T\\& -  \overline{\bm{a}}_i \bm{M}(\bm{S}_i^T\bm{S}_i)^{-1}\bm{M}^T \bm{P}_{\bm{S}_i} \bm{a}_i^T\\&+\biggl.\bm{a}_i\bm{S}_i(\bm{S}_i^T\bm{S}_i)^{-1} \bm{S}^T_i\bm{M}(\bm{S}_i^T\bm{S}_i)^{-1}\bm{M}^T\bm{P}_{\bm{S}_i}\bm{a}_i^T\biggr)\\&\leq O(r\epsilon)+O(r\epsilon)-O\left(\frac{r\eta^2}{k}\right)+O(r\epsilon)\\&=
-O\left(\frac{r\eta^2}{k}\right),
\end{align*}
as long as $\epsilon\leq O(\frac{\eta^2}{k})$. Therefore, for $\|\nabla c(\bm{S})\|\leq \epsilon $ and $\bm{S}$ at least $\eta = O(\sqrt{\epsilon})$ away from a global minimum, we have $\lambda_{min}(\nabla^2 c(\bm{S}))\leq -O(\frac{r\epsilon}{k})$. Thus, the desired statement

``For all $\bm{S}$ such that $\|\nabla c(\bm{S})\|\leq \epsilon$ and $\bm{S}$ more than $\eta$-away from a \emph{global} minimum, we have that $\lambda_{\text{min}}(\nabla^2f(\bm{x}))\leq -\zeta$ for some suitable $\zeta$."

is true for $\eta = O(\sqrt{\epsilon})$ and $\zeta = O(\frac{r\epsilon}{k})$, with probability $1-O(\frac{1}{n})$.

Thus, all local minima are global minima, and in particular the function $c(\bm{S})$ satisfies a $(\epsilon, O(\frac{r\epsilon}{k}), O(\sqrt{\epsilon}))$-strict saddle for sufficiently small $\epsilon$. 
\vfill
\pagebreak

\section{Proof of Technical Lemmas}
\begin{lem}
	\label{lem:hoeffding_general}
	Let $X_1,\cdots,X_n$ be independent (but not necessarily identically distributed) random variables which satisfy the following:
	\begin{equation*}
	\md{P}\left(|X_i-a_i|\geq \sigma_i \sqrt{\log(\frac{1}{\epsilon})}\right)\leq \epsilon .
	\end{equation*}
	Then we have 
	\begin{equation}
	\md{P}\left(\left|\frac{\sum_{i=1}^n X_i-a_i}{n}\right|\geq \frac{2\sqrt{\sum_{i=1}^n \sigma_i^2}}{n}\sqrt{\log(\frac{1}{4\epsilon})}\right)\leq \epsilon.
	\end{equation}
\end{lem}
\begin{proof}
	See \cite{bertsimas2018interpretable}.
\end{proof}

\begin{lem}
	\label{lem:matrix_inv_ineq}
	Assume that $\bm{S}$ satisfies Assumption \ref{ass:Vrestrict}. Then we have 
	\begin{equation}
	\md{P}\left(\left\|\left(\frac{\bm{S}^T\bm{S}}{m}\right)^{-1}-\left(\frac{\bm{S}_i^T\bm{S}_i}{l}\right)^{-1}\right\|\leq \sqrt{\frac{D\log(\frac{1}{\epsilon})}{l}}\right)\geq 1- \epsilon. \label{eq:matrix_inv_ineq} 
	\end{equation}
\end{lem}
\begin{proof}
	This follows immediately from Lemma \ref{lem:matrix_ineq} and the following lemma from matrix perturbation theory (for proof, see e.g. \cite{stewart1990matrix}):
	\begin{lem}
		\label{lem:matrix_inv}
		Let $\bm{A},\bm{B}$ be invertible matrices and let $\bm{B}=\bm{A}+\bm{\Delta}$. Then, we have the following bound:
		\begin{equation}
		\|\bm{A}^{-1}-\bm{B}^{-1}\|\leq \|\bm{A}^{-1}\|\|\bm{B}^{-1}\|\|\bm{\Delta}\|.
		\end{equation}
	\end{lem}
\end{proof}
\begin{lem}
	\label{lem:matrix_ineq}
	\begin{equation}
	\md{P}\left(\left\|\frac{\bm{S}^T\bm{S}}{m}-\frac{\bm{S}_i^T\bm{S}_i}{l}\right\|\leq \sqrt{\frac{C\log(\frac{1}{\epsilon})}{l}}\right)\geq 1- \epsilon. \label{eq:matrix_dev_bound3} 
	\end{equation}
\end{lem}
To prove this, we 
first introduce a matrix analog of the well-known Chernoff bound, the proof of which can be found in \cite{tropp2012user}:
\begin{lem}
	\label{lem:matrix_chernoff}
	Let $\mathcal{X} \in \md{R}^{k \times k}$ be a finite set of positive-semidefinite matrices, and suppose that
	\[\max_{\bm{X} \in \mathcal{X}} \lambda_{\max}(\bm{X})\leq D,\]
	where $\lambda_{\min}/\lambda_{\max}$ is the minimum/maximum eigenvalue function. Sample $\{\bm{X}_1,\cdots, \bm{X}_\ell\}$ uniformly at random without replacement. Compute:
	\[\mu_{\min}:=\ell \cdot \lambda_{\min}(\md{E} \bm{X}_1)\qquad \mu_{\max}:=\ell \cdot \lambda_{\max}(\md{E} \bm{X}_1).\]
	Then:
	\begin{align*}
	\md{P}\left\{\lambda_{\min} \left(\sum_j \bm{X}_j\right)\leq (1-\delta) \mu_{\min} \right\}&\leq k \cdot \exp\left(\frac{-\delta^2\mu_{\min}}{4D}\right) \quad \text{for } \delta \in [0,1),\\
	\md{P}\left\{\lambda_{\max} \left(\sum_j \bm{X}_j\right)\leq (1+\delta) \mu_{\max} \right\}&\leq k \cdot \exp\left(\frac{-\delta^2\mu_{\max}}{4D}\right) \quad \text{for } \delta \geq 0.       
	\end{align*}
\end{lem}
Now we  proceed with the proof.
\proof{(Lemma \ref{lem:matrix_ineq})} First, let us write $L$ as the set of indices $j$ such that $(\bm{W}_i)_{jj}=1$. Then, we  decompose $\bm{S}=\bm{Q}\bm{R}$ in a reduced $\bm{Q}\bm{R}$ factorization, where $\bm{Q}^T\bm{Q}=m\bm{I}$ (so that $\|\bm{R}\|=O(1)$). Then define $\bm{Q}_L$ as the $|L|\times k$ submatrix of $\bm{Q}$ formed with the rows in $L$. Then we can see that
\[\bm{S}^T\bm{S}=\bm{R}^T\bm{Q}^T\bm{Q}\bm{R},\qquad \bm{S}_i^T\bm{S}_i=\bm{R}^T\bm{Q}_L^T\bm{Q}_L\bm{R}.\]
Now let us decompose the inner parts, which can be written using the rows of $\bm{Q}$, $\bm{q}_i \in \md{R}^{1 \times k}$:
\begin{align*}
\bm{Q}^T\bm{Q}&=\sum_{i=1}^m \bm{q}_i^T\bm{q}_i,\\
\bm{Q}^T_L\bm{Q}_L&=\sum_{i\in L} \bm{q}_i^T\bm{q}_i,
\end{align*}
where $\bm{q}_i^T\bm{q}_i \in \md{R}^{k \times k}$ rank-one positive semi-definite matrices. Therefore, we can take $\bm{Q}^T_L\bm{Q}_L$ as a random sample of size $l$ from the set $\mathcal{X}=\{\bm{q}_i^T\bm{q}_i\}_{i=1,\cdots,m}$, which satisfies the conditions in Lemma \ref{lem:matrix_chernoff} with $D=O(1)$. Furthermore, with $\mathcal{X}$, we observe that we have $\md{E} \bm{X}_1=\frac{\bm{Q}^T\bm{Q}}{m}=\bm{I}_k$, so we have
\[\lambda_{\min}(\md{E} \bm{X}_1)=\lambda_{\max}(\md{E} \bm{X}_1)=1.\]
Therefore, we apply Lemma \ref{lem:matrix_chernoff} to $\bm{Q}_L^T\bm{Q}_L$ and have that
\begin{align*}
\md{P}\left\{\lambda_{\min} \left(\bm{Q}_L^T\bm{Q}_L\right)\leq (1-\delta)l\right\}&\leq \exp\left(\frac{-\delta^2l}{D'}\right),\\
\md{P}\left\{\lambda_{\max} \left(\bm{Q}_L^T\bm{Q}_L\right)\geq (1+\delta)l\right\}&\leq \exp\left(\frac{-\delta^2l}{D'}\right),
\end{align*}
where we set $D=\frac{D'}{4}$ with $D'=O(1)$. Some rearrangement gives:
\begin{equation}
\label{eq:matrix_eig_ineq}
\md{P}\left\{\lambda_{\min} \left(\frac{\bm{Q}_L^T\bm{Q}_L}{l}\right)\geq 1-\sqrt{\frac{D'\log\left(\frac{2}{\epsilon}\right)}{l}} \;\;  \text{and} \;\; \lambda_{\max} \left(\frac{\bm{Q}_L^T\bm{Q}_L}{l}\right)\leq 1+\sqrt{\frac{D'\log\left(\frac{2}{\epsilon}\right)}{l}}\right\}\geq 1 - \epsilon,
\end{equation}
Now since $\frac{\bm{Q}^T\bm{Q}}{m}=\bm{I}_k$, we have
\begin{equation}
\label{eq:matrix_eig_ineq2}
\lambda_{\min}\left(\frac{\bm{Q}^T\bm{Q}}{m}\right)=\lambda_{\max}\left(\frac{\bm{Q}^T\bm{Q}}{m}\right)=1.
\end{equation}
Combining equation (\ref{eq:matrix_eig_ineq2})
and (\ref{eq:matrix_eig_ineq}) gives:
\begin{equation}
\md{P}\left\{\left\|\frac{\bm{Q}_L^T\bm{Q}_L}{l}-\frac{\bm{Q}^T\bm{Q}}{m}\right\|\leq \sqrt{\frac{D'\log\left(\frac{2}{\epsilon}\right)}{l}}\right\}\geq 1 - \epsilon.
\end{equation}
Then, we have 
\begin{equation}
\md{P}\left\{\left\|\frac{\bm{R}^T\bm{Q}_L^T\bm{Q}_L\bm{R}}{l}-\frac{\bm{R}^T\bm{Q}^T\bm{Q}\bm{R}}{m}\right\|\leq \|\bm{R}\|^2\sqrt{\frac{D'\log\left(\frac{2}{\epsilon}\right)}{l}}\right\}\geq 1 - \epsilon.
\end{equation}
Taking $C=D'\|\bm{R}\|^4\log(2)$ gives the required result.
\endproof

\section{List of Features for Netflix Data}
\label{app:featurelist}
\begin{itemize}
    \item 24 Indicator Variables for Genres: Action, Adventure, Animation, Biography, Comedy, Crime, Documentary, Drama, Family, Fantasy, Film Noir, History, Horror, Music, Musical, Mystery, Romance, Sci-Fi, Short, Sport, Superhero, Thriller, War, Western
    \item 5 Indicator Variables for Release Date: Within last 10 years, Between 10-20 years, Between 20-30 years, Between 30-40 years, Between 40-50 Years
    \item 6 Indicator Variables for Top Actors/Actresses defined by their Influence Score at time of release: Top 100 Actors, Top 100 Actresses, Top 250 Actors, Top 250 Actresses, Top 1000 Actors, Top 1000 Actresses
    \item IMDB Rating
    \item Number of Reviews
    \item Total Production Budget
    \item Total Runtime
    \item Total Box Office Revenue
    \item Indicator Variable for whether it is US produced
    \item 11 Indicator Variables for Month of Year Released (January removed to prevent multicollinearity)
    \item Number of Original Music Score
    \item Number of Male Actors
    \item Number of Female Factors
    \item 3 Indicator Variables for Film Language: English, French, Japanese
    \item Constant 
\end{itemize}
\end{document}